\documentclass[journal]{IEEEtran}

\usepackage{graphics}
\usepackage{pdfpages}
\usepackage{cite}
\usepackage{amsmath}
\usepackage{amsfonts}
\usepackage{bm}
\usepackage{comment}
\usepackage{booktabs}
\usepackage{multirow}
\usepackage{mathtools}
\usepackage{amssymb}
\usepackage{pifont}
\usepackage{float}
\usepackage{epsfig}
\usepackage{array}
\usepackage{siunitx}
\usepackage{threeparttable}
\usepackage[table]{xcolor}
\definecolor{ForceHighlight}{RGB}{220,235,247}
\definecolor{MomentHighlight}{RGB}{248,224,224}
\definecolor{OpMid}{RGB}{255,255,255}
\definecolor{BestShade}{RGB}{226,239,218}
\definecolor{WorstShade}{RGB}{255,242,204}
\definecolor{SelectedKeyShade}{RGB}{210,232,199}
\definecolor{SelectedWeakShade}{RGB}{255,242,204}
\newcommand{\bestval}[1]{\begingroup\setlength{\fboxsep}{1pt}\colorbox{BestShade}{\ensuremath{#1}}\endgroup}
\newcommand{\worstval}[1]{\begingroup\setlength{\fboxsep}{1pt}\colorbox{WorstShade}{\ensuremath{#1}}\endgroup}
\newcommand{\keyval}[1]{\begingroup\setlength{\fboxsep}{1pt}\colorbox{SelectedKeyShade}{\ensuremath{#1}}\endgroup}
\newcommand{\weakval}[1]{\begingroup\setlength{\fboxsep}{1pt}\colorbox{SelectedWeakShade}{\ensuremath{#1}}\endgroup}
\usepackage{soul}
\usepackage{algorithmicx}
\usepackage[ruled]{algorithm}
\usepackage[noend]{algpseudocode}
\usepackage{url}
\usepackage{subcaption}
\usepackage{booktabs}
\newcommand{\etal}{\textit{et al}.}

\begin{document}

\title{Planning a Shared Modular Fixture Layout Across Robotic Disassembly Stages}

\author{Haohui~Pan, Takuya~Kiyokawa, and~Kensuke~Harada%
\thanks{Corresponding author: Haohui Pan.}%
\thanks{H. Pan and T. Kiyokawa are with the Department of Systems Innovation, Graduate School of Engineering Science, The University of Osaka, 1-3 Machikaneyama, Toyonaka, Osaka, Japan.}%
\thanks{K. Harada is with the Department of Systems Innovation, Graduate School of Engineering Science, The University of Osaka, 1-3 Machikaneyama, Toyonaka, Osaka, Japan, and also with the Industrial Cyber-physical Systems Research Center, The National Institute of Advanced Industrial Science and Technology (AIST), 2-3-26 Aomi, Koto-ku, Tokyo, Japan.}%
}

\markboth{IEEE Transactions on Automation Science and Engineering}%
{Pan \MakeLowercase{\etal}: Shared Modular Fixture Layout Planning}

\maketitle

\begin{abstract}
Stable support remains challenging in robotic disassembly of irregularly shaped products. As components are progressively removed, the available support surfaces, mass distribution, and task loads change throughout the process. A fixture layout designed for one workpiece state may therefore become infeasible at later stages, motivating unified support planning over the complete disassembly sequence. This paper presents a modular vacuum-based fixturing system that plans one shared support configuration for the complete disassembly sequence of a screwdriver or shaver, allowing each sequence to proceed without fixture reconfiguration. To search the mixed continuous--discrete layout space under repeated cross-stage evaluation, a denoising diffusion probabilistic model generates physics-informed initial configurations that are refined through Bayesian optimization. Robotic screw and component-removal experiments verified the disassembly feasibility of the planned layouts, while 11 directional-load tests quantified their stability. Comparisons between the measured operational loads and directional responses yielded mean empirical stability margins of 66.9\% for the screwdriver and 81.6\% for the shaver. These results demonstrate that a product-specific shared layout can provide stable support throughout the tested robotic disassembly sequence.
\end{abstract}

\def\abstractname{Note to Practitioners}
\begin{abstract}
Fixtures used in automated disassembly are commonly designed for a particular
product shape or operation. For products with irregular surfaces, rigid
clamping may provide inadequate support or obstruct access to removable
components, while fixture adjustment between operations interrupts the
workflow. The proposed system is intended for robotic disassembly cells in
which the sequence is known in advance but the available support surfaces
change as components are removed. Given the object states, candidate support
regions, and expected operation loads, the method determines the locations and
models of modular vacuum supports that can remain unchanged throughout the
sequence. This enables an engineer to prepare one layout before disassembly
rather than reconfigure the fixture at every stage. Physical experiments show
that the resulting layouts accommodate the tested screw and successive
component-removal operations with positive empirical stability margins.
The current implementation assumes quasi-static operations, predefined object
states, and known or estimated task loads. End-effector alignment is performed
manually, and unexpected state changes do not yet trigger online fixture
replanning. Practical deployment would therefore benefit from
perception-based alignment, contact monitoring, and online replanning.

\end{abstract}


\IEEEpeerreviewmaketitle

\section{Introduction}

\IEEEPARstart{T}{he} recycling and reuse of engineering products increasingly
require disassembly to recover components and valuable materials. Component
removal changes the accessible surface, mass distribution, and loading of the
remaining assembly. A fixture suitable for one state may therefore obstruct a
later operation or lose support quality, motivating planning over the complete
sequence rather than a single static model.

Products such as hair dryers, electric drills, and shavers often have
irregular surfaces and asymmetric geometries that are difficult to constrain
with conventional rigid fixtures, especially as their geometry and support
conditions change during disassembly. Supporting these products requires
contact elements that accommodate local curvature without enclosing the
object or obstructing the operation space. A balloon
hand\footnote{https://convum.co.jp/products/en/other-en/sgb/} provides a
compliant contact surface that conforms to local geometry and uses vacuum
suction for retention. Multiple independently arranged balloon hands can
therefore provide localized support while preserving access for screw and
component-removal operations, and are adopted as the support elements in this
study.

In addition to geometric changes, individual disassembly operations impose
different task loads on the supported object. Screw removal, for example,
introduces explicit force and torque demands that must be considered when
evaluating fixture stability. The planned fixture layout must therefore remain
feasible under the operation loads prescribed at each stage.

\begin{figure}[!t]
    \centering
    \includegraphics[width=1\linewidth]{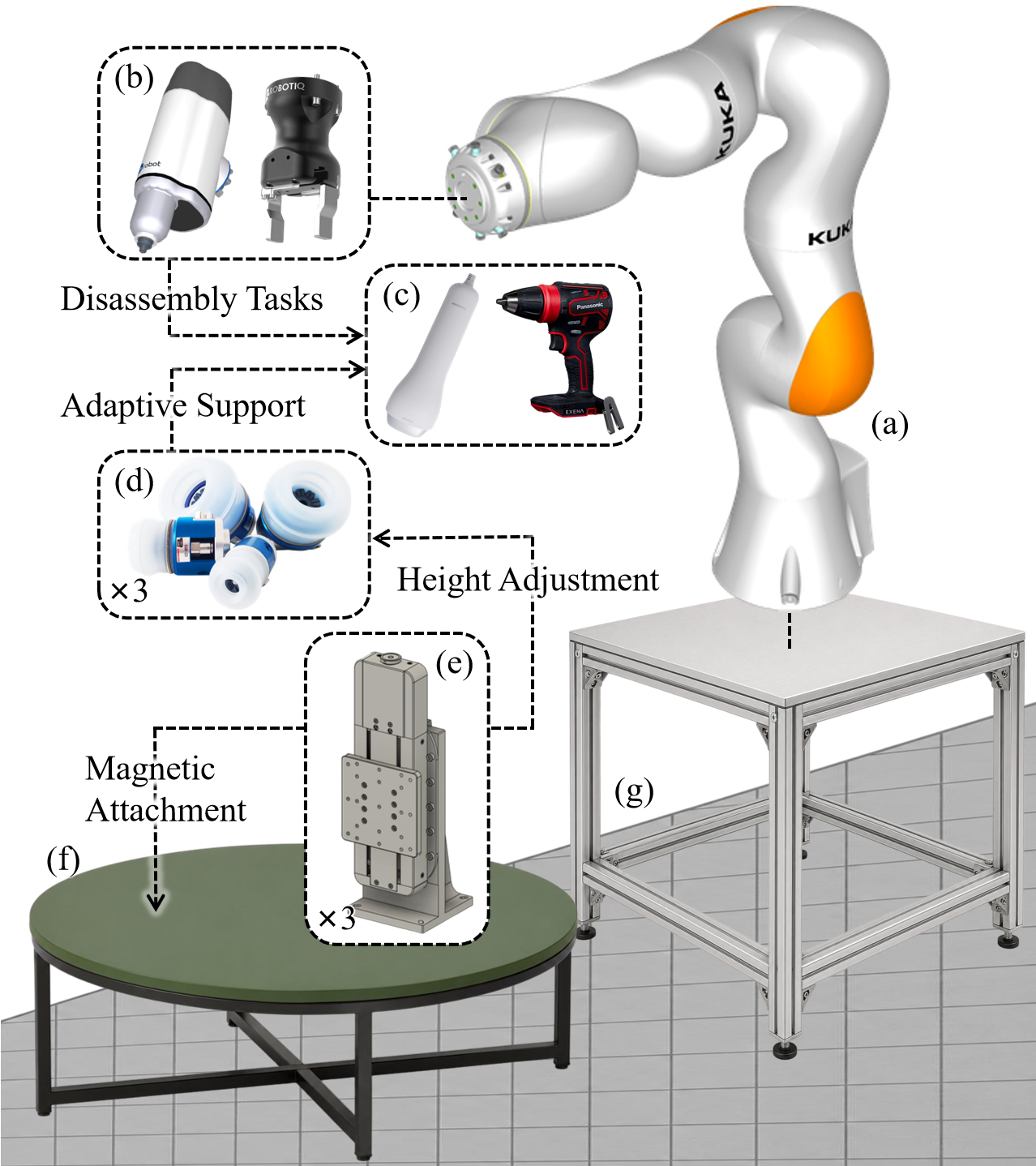}
    \caption{Robotic disassembly workspace. (a) KUKA LBR iiwa 14 manipulator; (b) OnRobot Screwdriver and Robotiq Hand-E gripper; (c) Panasonic screwdriver and Panasonic shaver; (d) three vacuum-based balloon hands; (e) three height-adjustment modules; (f) magnetic worktable; and (g) robot table.}
    \label{fig:workspace}
\end{figure}

To address this problem, this paper proposes a modular vacuum-based fixturing
system for products with irregular surfaces. As shown in
Fig.~\ref{fig:workspace}, the system integrates three balloon hands with
independent height-adjustment modules on a magnetic worktable and a KUKA LBR
iiwa 14 manipulator equipped with an OnRobot Screwdriver or a Robotiq Hand-E
gripper. A Panasonic screwdriver and shaver are used as the target objects.

Pan et al.~\cite{pan2026modular} applied modular vacuum-based fixturing to
screw-removal tasks in a single workpiece state and showed that three balloon
hands provided more stable support than two. Based on this finding, the present
study adopts the three-hand structure and extends the single-state evaluation
to complete disassembly sequences by planning a product-specific shared fixture
layout that remains unchanged across all stages. Retaining one layout
throughout the sequence avoids fixture reconfiguration after each
component-removal operation.

The main contributions are threefold. First, an adaptive compliant support
system is developed for products with irregular surfaces. Second, a
cross-stage support-layout optimization method is proposed to determine a
shared fixture layout for each target object under changing geometry and task
loads throughout its complete disassembly process. Third, the support stability of the proposed system
is qualitatively and quantitatively validated through robotic disassembly and directional-load
experiments.

\section{Related Work}

Planning stable support over a complete robotic disassembly sequence requires
three related capabilities: representing stage-dependent operations and task
loads, adapting the fixture to irregular and changing geometries, and searching
mixed-variable support configurations efficiently. Accordingly, this section
reviews disassembly planning and robotic execution, reconfigurable fixturing
and adaptive support, and generative design and Bayesian optimization.

\subsection{Disassembly Planning and Robotic Execution}

Sequence-level studies determine the order of component removal under process
objectives and constraints. Ren et al.~\cite{ren2020multiobjective} balanced
value recovery and energy consumption in a multiobjective formulation. Tian et
al.~\cite{tian2017disassembly} incorporated uncertain component quality and
varying operational cost into disassembly-sequence planning. Guo et
al.~\cite{guo2017dual} optimized sequences under multiple resource constraints
using a dual-objective formulation and scatter search. Wang et
al.~\cite{wang2021energy} developed an energy-efficient method for robotic
parallel disassembly. These studies optimize process-level decisions but
generally assume that suitable workpiece support remains available throughout
the resulting sequence.

Beyond process-level objectives, other studies consider evolving product states
and execution uncertainty. Bedeoui et al.~\cite{bedeoui2019assembly} generated
assembly sequences for heavy machines using a stability criterion based on the
evolving assembly state. Laili et al.~\cite{laili2021robotic} introduced backup actions into
robotic disassembly-sequence planning to accommodate execution failures and
uncertain product conditions. Peng et al.~\cite{peng2025dynamic} used
multi-agent deep reinforcement learning for dynamic human--robot collaborative
disassembly planning. Although these methods account for state changes or
uncertainty, their planning variables concern actions and recovery strategies
rather than a fixture layout shared across changing object states.

At the operation level, other studies focus on sensing and interaction
modeling. DiFilippo and Jouaneh~\cite{difilippo2017system}
combined force and vision sensing for automated screw removal from laptops.
Tao et al.~\cite{tao2026robotic} modeled forces and moments in compliant
robotic disassembly under an uncertain incomplete state. These studies improve individual operations
but treat fixture support as given. The present study instead connects
sequence-level planning and execution by requiring one fixture layout to remain
feasible under the geometry and task loads of every predefined stage.

\begin{figure*}[!t]
    \centering
    \includegraphics[width=1\linewidth]{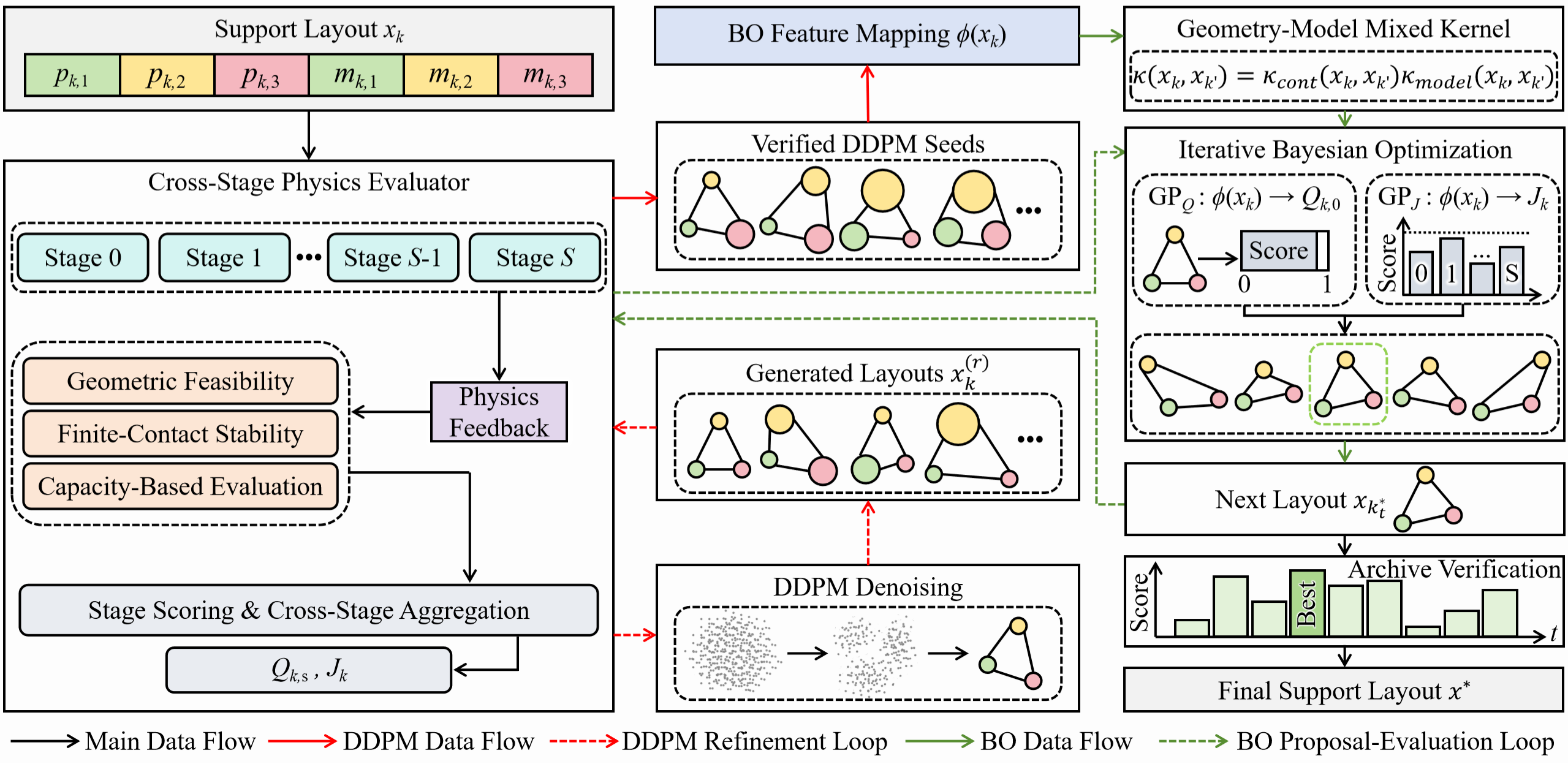}
\caption{Physics-guided support-planning framework integrating cross-stage evaluation, DDPM refinement, and Bayesian optimization.}
    \label{fig:system}
\end{figure*}

\subsection{Reconfigurable Fixturing and Adaptive Support}

Complementing disassembly planning, early fixture research established
computer-aided representations and design methods. Whybrew and
Ngoi~\cite{whybrew1992computer} developed computer-aided modular-fixture
assembly, Wang et al.~\cite{wang2010computer} reviewed computer-aided fixture
design, Asada and By~\cite{asada1985kinematic} analyzed automatically reconfigurable
fixtures, and Xiong et al.~\cite{xiong2007algebraic} represented rigid complex
fixture systems algebraically and geometrically. These methods support fixture
construction but do not consider whether one layout remains feasible as the
object changes during disassembly.

Extending these foundations, Naing et al.~\cite{naing2000design} proposed
integrated tooling for jigless assembly, while Bi and
Zhang~\cite{bi2001flexible} reviewed flexible-fixture design and automation.
These architectures enable reusable tooling but do not plan support throughout
a known sequence of changing object states.

Building on these architectures, Chan et al.~\cite{chan1990reconfigurable}
developed a reconfigurable fixture for robotic assembly, Yu et
al.~\cite{yu2012conceptual} designed fixturing robots for sheet-metal assembly,
and Fathianathan et al.~\cite{fathianathan2007adaptive} developed adaptive
machining-fixture design. Bejlegaard et
al.~\cite{bejlegaard2018methodology} studied reconfigurable fixture
architectures; Park et al.~\cite{park2021assembly} and Shi et
al.~\cite{shi2021development} developed transformable pin-array fixtures;
Nelaturi et al.~\cite{nelaturi2014automated} automated configuration under
constraint and accessibility requirements; and de Leonardo et
al.~\cite{de2013swarmitfix} constructed fixtures using mobile modules. These
systems achieve adaptability through physical reconfiguration rather than one
layout retained across successive component-removal stages.

In parallel, fixture-layout optimization has addressed localization, force
closure, deformation, and contact-force distribution. Li and
Melkote~\cite{li2001optimal} considered workpiece dynamics, Kim and
Ding~\cite{kim2004optimal} optimized layouts across assembly stations, and Zhu
and Ding~\cite{zhu2009optimality} compared localization, force, and deformation
criteria. Pelinescu and Wang~\cite{pelinescu2002multi} formulated a
multiobjective layout problem, while Kaya~\cite{kaya2006machining}, Lu and
Zhao~\cite{lu2015fixture}, and Arzanpour et
al.~\cite{arzanpour2006flexible} optimized fixtures for machining, deformable
sheet metal, and automotive assembly, respectively. Liu et
al.~\cite{liu2024optimizing} and Meng et al.~\cite{meng2023intelligent} further
optimized layouts for compliant parts and multiple aircraft panels. These
methods address specified manufacturing states or workpieces rather than one
layout evaluated under stage-dependent disassembly geometries and task loads.

More recently, compliant and suction-based devices have enabled local
adaptation to object geometry. Kemmotsu et
al.~\cite{kemmotsu2024balloon} developed a balloon pin-array gripper that
adapts to object misalignment. Sakuma et al.~\cite{sakuma2023jamming}
developed a jamming-gripper-inspired soft jig for adaptive fixing and contact
perception. Lee et al.~\cite{lee2025grasp} formulated failure constraints for
reliable manipulation with multiple suction cups, and Guo et
al.~\cite{guo2026novel} developed self-sealing suction mechanisms. Hong et
al.~\cite{hong2026integrated} jointly optimized tolerances and layouts of
flexible fixtures, while Arroyo et al.~\cite{arroyo2026development} developed
a self-adjusting fixturing system. These devices improve local geometric
adaptability but do not couple compliance with sequence-wide fixture-layout
planning. The present study jointly plans the positions and models of
vacuum-based balloon hands for all stages of a predefined disassembly sequence.

\subsection{Generative Design and Bayesian Optimization}

Beyond physical fixture design, cross-stage planning requires efficient search
because each mixed-variable candidate undergoes expensive physical evaluation.
Weng et al.~\cite{weng2024dexdiffuser} proposed DexDiffuser to generate and
rank dexterous grasp candidates. Inoue et al.~\cite{inoue2023layoutdm}
developed a discrete diffusion model for controllable layout generation. Li et
al.~\cite{li2025expensive} integrated diffusion-based generation with BO for
expensive multiobjective problems. These studies demonstrate structured
candidate generation but do not address support layouts whose feasibility must
persist across changing object states.

As a complementary approach, BO supports data-efficient search under expensive
or hybrid evaluations. Greenhill et
al.~\cite{greenhill2020bayesian} reviewed BO for adaptive experimental design
with expensive black-box evaluations. Zhang et
al.~\cite{zhang2020bayesian} constructed a BO framework for quantitative and
qualitative variables. These methods improve
data-efficient search but do not use physics-refined generative initialization
to search for a fixture layout shared across a complete disassembly sequence.
The present framework therefore initializes BO with physics-refined DDPM
samples and uses a geometry--model mixed kernel for the mixed
continuous--discrete support configurations.

\begin{figure*}[t]
    \centering
    \includegraphics[width=1\linewidth]{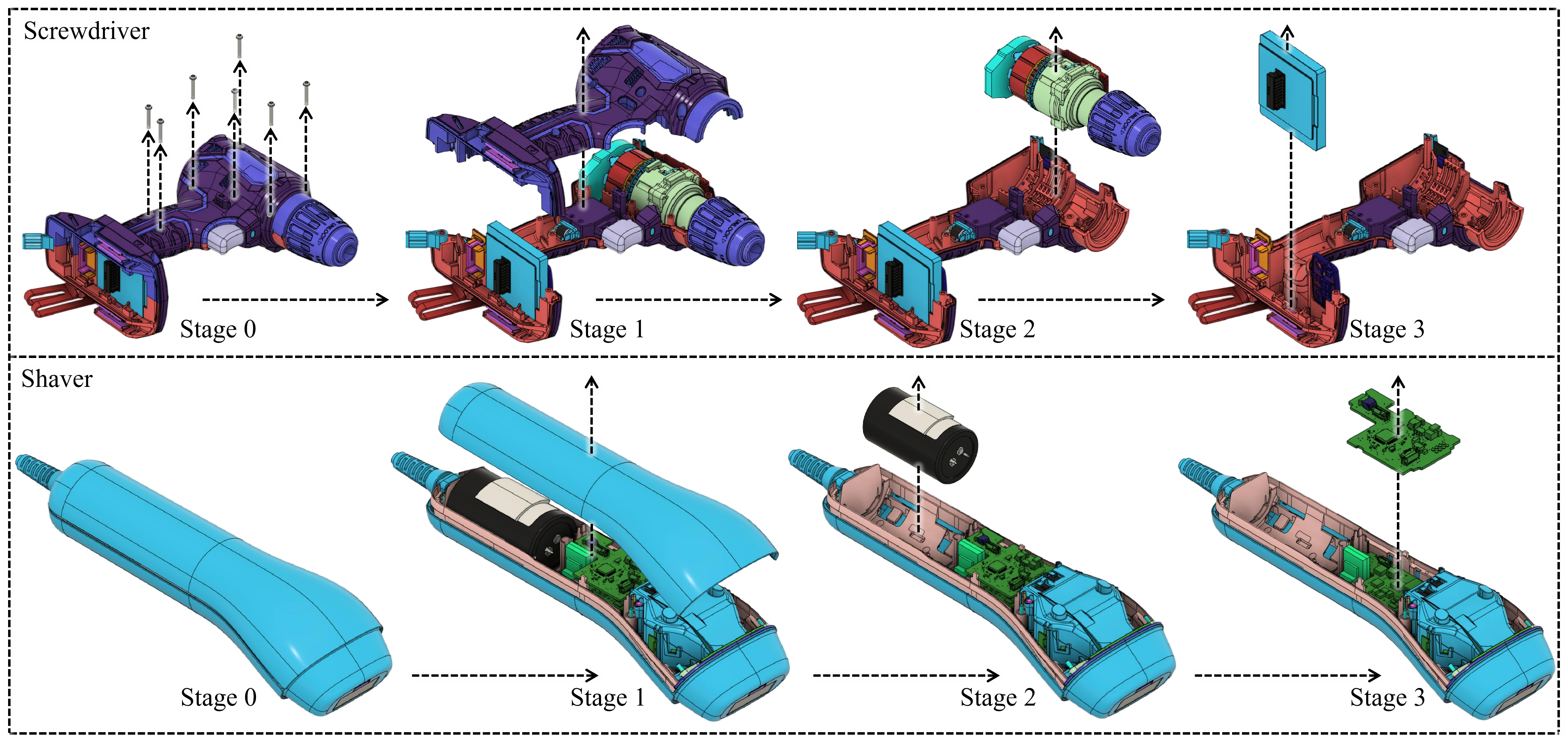}
    \caption{Predefined disassembly sequences for the screwdriver (top) and shaver (bottom). Stages 0--3 show the intact product and three successive component-removal states; arrows identify the component removed between adjacent stages.}
    \label{fig:stage-setting}
\end{figure*}

\section{Proposed Method}

This section presents the proposed support-planning framework. It first defines the system assumptions and mathematical model, and then describes the DDPM-based initialization and BO-based layout refinement.

\subsection{System Overview and Assumptions}

Fig.~\ref{fig:system} summarizes the proposed framework. For the \(k\)-th candidate, \(i\in\{1,2,3\}\) indexes the three labeled balloon-hand modules. The position and model assigned to the \(i\)-th module are denoted by \(p_{k,i}=[p_{k,i,x},p_{k,i,y},p_{k,i,z}]^T\in\mathbb{R}^3\) and \(m_{k,i}\), respectively. This indexing is retained throughout the candidate representation, physical evaluation, and BO feature mapping. The candidate support configuration is defined as
\begin{equation}
\begin{aligned}
x_k=\big[
&p_{k,1},p_{k,2},p_{k,3},
m_{k,1},m_{k,2},m_{k,3}
\big],
\end{aligned}
\label{eq:support_configuration}
\end{equation}
where \(m_{k,i}\in\mathcal{M}_{\mathrm{model}}\) is a discrete variable drawn
from the finite set of available balloon-hand models. Let
\(s\in\mathcal{S}_{\mathrm{all}}=\{0,\ldots,S\}\) index the predefined
disassembly stages.

The cross-stage physics evaluator maps each \(x_k\) to every predefined stage
and returns the stage qualities \(Q_{k,s}\) and sequence objective \(J_k\).
Physics-evaluated configurations guide DDPM refinement, and the verified final
outputs initialize BO. Two Gaussian-process surrogates then guide iterative
candidate selection using the geometry--model mixed kernel. After the BO
budget is exhausted, the deduplicated archive is verified across all stages,
and \(x^\ast\) denotes the all-stage-feasible layout with the highest \(J_k\).

The disassembly sequence is predefined. As shown in
Fig.~\ref{fig:stage-setting}, each object has four states beginning with the
intact product. The screwdriver sequence removes the upper housing, motor, and
circuit board after the required screw operations, while the shaver sequence
removes the upper cover, battery, and circuit board. At each stage, the mesh,
center of gravity (CoG) \(C_s\), mass, candidate regions, and operation loads
are known or estimated. The nominal pose keeps the target screw axes aligned
with the global upward direction \(\hat z\) and remains fixed during support
generation and evaluation. Minor physical alignment deviations are treated as
deployment disturbances.

The compliant balloon-hand surface conforms to local irregular geometry,
enlarging the effective contact region without exact fixture--object matching.
Because deformation, contact area, vacuum pressure, and friction are strongly
coupled, their full nonlinear mechanics are impractical for repeated
cross-stage optimization. The physics-based quasi-static model therefore
captures finite-contact tipping resistance, tensile suction capacity, and
axial torsional capacity. Each active balloon hand supplies a vertical support
force at its center and bounded torsional resistance about its axis. Repeated
robotic disassembly and directional-load experiments subsequently evaluate
this model's applicability to support planning.
\subsection{Mathematical Model}

This subsection defines the physical model used to evaluate the feasibility and quality of a balloon-hand support configuration.

\subsubsection{Disassembly Stage and Candidate Representation}

At stage \(s\), the object state is represented by the mesh \(\mathcal{M}_s\), mass \(M_s\), CoG \(C_s\in\mathbb{R}^3\), and the operation-dependent force, torque, and application point. These quantities vary with the object state and provide the known inputs to the stage evaluator.

\begin{figure}[!t]
    \centering
    \includegraphics[width=\linewidth]{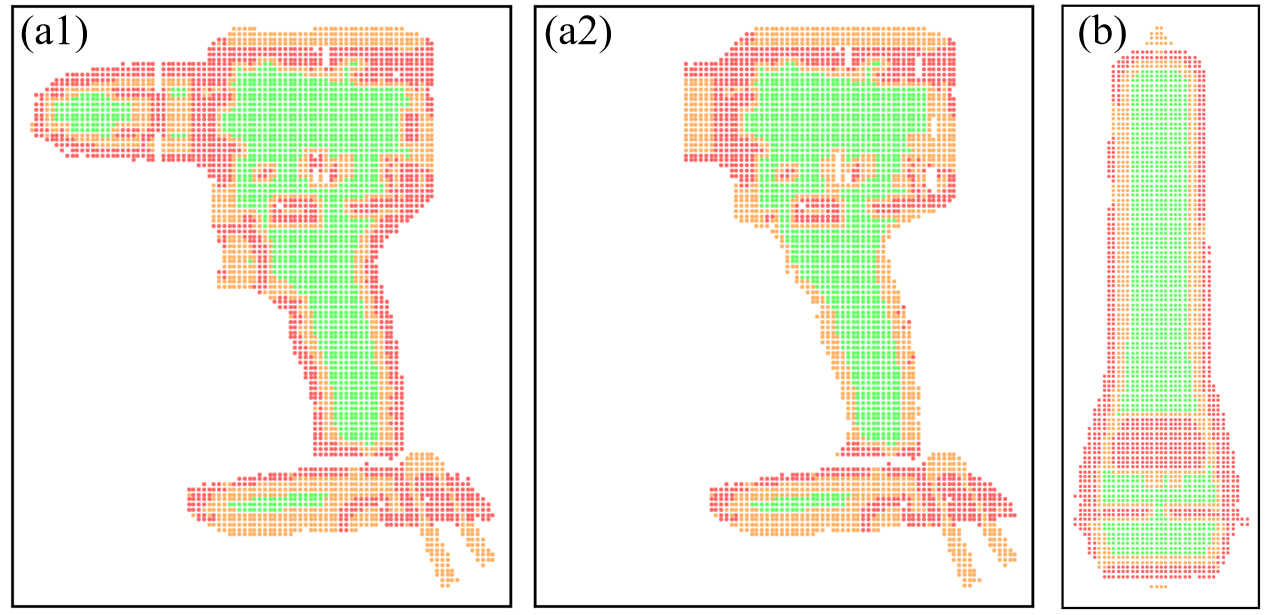}
\caption{Candidate-point classifications for the predefined object states: (a1) screwdriver, stages 0--1; (a2) screwdriver, stages 2--3; and (b) shaver, stages 0--3.}
    \label{fig:rgo}
\end{figure}

A stage-dependent surface set \(\mathcal{P}_s\) is sampled from
\(\mathcal{M}_s\) and partitioned into the risk region \(\mathcal{R}_s\),
feasible suction candidates \(\mathcal{G}_s\), and boundary region
\(\mathcal{E}_s\). A sampled point is
\(u_a=[u_{a,x},u_{a,y},z_a]^T\), with planar projection
\(u_{a,xy}=[u_{a,x},u_{a,y}]^T\). The risk region contains points having a
neighbor within planar radius \(r_{xy}\) whose height differs by more than
\(z_{\mathrm{th}}\):
\begin{equation}
\begin{aligned}
\mathcal{R}_s
={}&
\bigl\{u_a\in\mathcal{P}_s\mid
\exists u_b\in\mathcal{P}_s:\,
\\[-1mm]
&0<\|u_{b,xy}-u_{a,xy}\|_2\leq r_{xy},\,
|z_b-z_a|>z_{\mathrm{th}}\bigr\},
\end{aligned}
\label{eq:red_point_set}
\end{equation}
Using the local surface-coverage ratio \(\rho_a\in[0,1]\) and its threshold
\(\rho_{\mathrm{cov}}\), the remaining sets are
\begin{equation}
\begin{aligned}
\mathcal{G}_s
&=\{u_a\in\mathcal{P}_s\setminus\mathcal{R}_s\mid
\rho_a\geq\rho_{\mathrm{cov}}\},\\
\mathcal{E}_s
&=\{u_a\in\mathcal{P}_s\setminus\mathcal{R}_s\mid
\rho_a<\rho_{\mathrm{cov}}\}.
\end{aligned}
\label{eq:green_orange_point_sets}
\end{equation}
Fig.~\ref{fig:rgo} shows the resulting classifications, where red, green, and
orange denote \(\mathcal{R}_s\), \(\mathcal{G}_s\), and \(\mathcal{E}_s\),
respectively.

The canonical support configuration is expressed in the common object coordinate frame. For
any \(q=[q_x,q_y,q_z]^T\), let \(q_{xy}=[q_x,q_y]^T\) denote its \(XY\)-plane
projection. At stage \(s\), each center is mapped in the \(XY\) plane to the
nearest point of \(\mathcal{G}_s\) within \(d_{\mathrm{map}}=1\) mm, while its
height is taken from that stage candidate:
\begin{equation}
\begin{aligned}
p_{k,s,i}
&=\Pi_s(p_{k,i}),\\
\mathcal{A}_{k,s}
&=\left\{
i\ \middle|\
d_{\mathcal G_s}(p_{k,i,xy})\leq d_{\mathrm{map}},
\ \operatorname{cov}(p_{k,s,i})\geq0.90
\right\}.
\end{aligned}
\label{eq:stage_resolved_contact}
\end{equation}
Here, \(\Pi_s\) is the stage-dependent nearest-candidate mapping described
above, \(d_{\mathcal G_s}(q)=\min_{u\in\mathcal G_s}\|q-u_{xy}\|_2\) for
\(q\in\mathbb{R}^2\) is the planar point-to-set distance, and
\(\operatorname{cov}(\cdot)\) is the supported fraction of the suction
perimeter. The active set \(\mathcal{A}_{k,s}\) captures contacts lost as
components are removed. Configurations retaining two or three contacts remain
subject to physical evaluation, and their stage quality and feasibility enter
the cross-stage optimization.

Four balloon-hand models are considered, as summarized in Table~\ref{tab:sgb_models}. For each model \(m\in\mathcal{M}_{\mathrm{model}}\), the lookup functions \(R_{\mathrm{suc}}(m)\), \(R_{\mathrm{phy}}(m)\), \(F_{\mathrm{suc}}(m)\), and \(T_{\mathrm{suc}}(m)\) return its suction coverage radius, physical collision radius, maximum suction force, and torsional resistance limit, respectively.

\begin{table}[t]
\centering
\caption{Physical parameters of the balloon-hand models.}
\label{tab:sgb_models}
\footnotesize
\setlength{\tabcolsep}{4pt}
\renewcommand{\arraystretch}{1.18}
\begin{tabular*}{\columnwidth}{@{\extracolsep{\fill}}c c c c c}
\toprule
Model & \(R_{\mathrm{suc}}\) (mm) & \(R_{\mathrm{phy}}\) (mm) & \(F_{\mathrm{suc}}\) (N) & \(T_{\mathrm{suc}}\) (N\(\cdot\)m) \\
\midrule
SGB10 & 5.0  & 15.0 & 7.5  & 0.06 \\
SGB20 & 10.0 & 21.5 & 23.0 & 0.11 \\
SGB30 & 15.0 & 33.0 & 57.0 & 0.17 \\
SGB40 & 20.0 & 37.5 & 85.0 & 0.23 \\
\bottomrule
\end{tabular*}
\end{table}

Accordingly, for the \(i\)-th balloon hand in candidate \(x_k\), let
\(R_{\mathrm{suc},k,i}=R_{\mathrm{suc}}(m_{k,i})\) and
\(R_{\mathrm{phy},k,i}=R_{\mathrm{phy}}(m_{k,i})\). The former is used to
evaluate surface coverage and finite-contact stability, whereas the latter is
used for physical collision checking.

\subsubsection{Geometric Feasibility Constraints}

Geometric feasibility combines the active-contact test in Eq.~\eqref{eq:stage_resolved_contact}, inter-support collision, and boundary margin. For active balloon hands \(i\) and \(j\), the model-dependent spacing constraint is
\begin{equation}
\begin{aligned}
\|p_{k,s,i}-p_{k,s,j}\|_2
&\geq R_{\mathrm{phy},k,i}+R_{\mathrm{phy},k,j}+d_{\mathrm{gap}},\\
&i,j\in\mathcal{A}_{k,s},\qquad i<j,
\end{aligned}
\label{eq:collision_constraint}
\end{equation}
where \(d_{\mathrm{gap}}\) is the design gap.

The boundary margin
\(d_{\mathrm{edge},k,s,i}=\min_{e\in\mathcal E_s}
\|p_{k,s,i}-e\|_2\) contributes to the stage quality and BO features.

\subsubsection{Finite-Contact Stability and Force--Torque Capacity}

Each active suction region is represented by a circle centered at
\(p_{k,s,i,xy}\) with radius \(R_{\mathrm{suc},k,i}\). The finite-contact
support region \(\mathcal{H}_{k,s}\) is formed by the exact outer common
tangents and contact chords of these circles. Stability requires
\(C_{s,xy}\in\mathcal{H}_{k,s}\) with the configured minimum tipping margin.

For one operation at stage \(s\), let
\(F_{\mathrm{op}},\tau_{\mathrm{op}}\in\mathbb{R}^3\) and
\(p_{\mathrm{op}}\in\mathbb{R}^3\) denote the task force, direct task torque,
and application point, respectively. Let \(g_0\) be gravitational acceleration
and \(\hat d_g\in\mathbb{R}^3\) the unit gravity direction in the object frame.
Under the prescribed screw-operation pose, the direct task torque is aligned
with the balloon-hand axis and is represented as
\(\tau_{\mathrm{op}}=T_{\mathrm{op}}\hat z\), where \(T_{\mathrm{op}}\) is
its signed magnitude.
The external wrench \(w_s^{\mathrm{ext}}\in\mathbb{R}^6\), expressed about the
CoG, is
\begin{equation}
w_s^{\mathrm{ext}}
=
\begin{bmatrix}
M_sg_0\hat d_g+F_{\mathrm{op}}\\
(p_{\mathrm{op}}-C_s)\times F_{\mathrm{op}}+\tau_{\mathrm{op}}
\end{bmatrix}.
\label{eq:external_wrench}
\end{equation}
When no screw-removal operation is required at a stage, \(F_{\mathrm{op}}\) and \(\tau_{\mathrm{op}}\) are set to zero, yielding the gravity-only case. Multiple screw-removal operations at one stage use the same formulation with their respective input data and are evaluated independently.

The balloon-hand axes remain aligned with
\(\hat z=[0,0,1]^T\). With \(n_{k,s}=|\mathcal A_{k,s}|\), let
\(A^F_{k,s}\in\mathbb R^{6\times n_{k,s}}\) be the stage-specific wrench mapping
whose \(i\)-th column is
\([\hat z^T,((p_{k,s,i}-C_s)\times\hat z)^T]^T\) for
\(i\in\mathcal A_{k,s}\). The center-force vector
\(f=[f_i]_{i\in\mathcal A_{k,s}}\in\mathbb{R}^{n_{k,s}}\) and direct torsional
moment vector
\(\mu=[\mu_i]_{i\in\mathcal A_{k,s}}\in\mathbb{R}^{n_{k,s}}\) are estimated
separately:
\begin{equation}
\begin{aligned}
f^*_{k,s}
&=\arg\min_f
\left\|A^F_{k,s}f+
\begin{bmatrix}
M_sg_0\hat d_g+F_{\mathrm{op}}\\
(p_{\mathrm{op}}-C_s)\times F_{\mathrm{op}}
\end{bmatrix}\right\|_2^2,\\
\mu^*_{k,s}
&=\arg\min_\mu
\left\|\sum_{i\in\mathcal{A}_{k,s}}\mu_i\hat z+
\tau_{\mathrm{op}}\right\|_2^2.
\end{aligned}
\label{eq:force_torque_demand}
\end{equation}
When a least-squares problem admits multiple minimizers, the
minimum-Euclidean-norm solution is used to obtain a deterministic allocation
of the demands among the active contacts.
The optimizers \(f^*_{k,s}\) and \(\mu^*_{k,s}\) collect the corresponding
per-contact demands \(f^*_{k,s,i}\) and \(\mu^*_{k,s,i}\). These demands must
satisfy
\begin{equation}
\begin{aligned}
f^*_{k,s,i}
&\geq-\frac{F_{\mathrm{suc}}(m_{k,i})}{\alpha_{\mathrm{sf}}},\\
|\mu^*_{k,s,i}|
&\leq T_{\mathrm{suc}}(m_{k,i}),
\qquad i\in\mathcal{A}_{k,s}.
\end{aligned}
\label{eq:force_torque_capacity}
\end{equation}
Here, \(\alpha_{\mathrm{sf}}>0\) is the suction-force safety factor. The negative
force bound follows from the sign convention: loading opposite to \(\hat z\)
represents the tensile demand resisted by suction.
The least-squares solution estimates the per-contact demands under the adopted
vertical contact-force and axial torsional-resistance representation. Load
components that cannot be represented by these contact actions remain in the
least-squares residual, which is monitored as a model-consistency diagnostic
rather than used as a hard feasibility criterion. Feasibility is determined
by the modeled physical failure criteria: finite-contact tipping, tensile
suction-capacity violation, and axial torsional-capacity violation.

\subsubsection{Support Quality Score}

Each stage evaluation returns the hard-feasibility indicator \(h_{k,s}\in\{0,1\}\) and stage quality \(Q_{k,s}\in[0,1]\). The same score is used for DDPM feedback, BO, and archive verification.

The score combines the normalized force-demand and risk-region penalties
\(I_{\mathrm{force},k,s}\) and \(P_{\mathrm{red},k,s}\), boundary-margin and
finite-contact-area rewards \(U_{\mathrm{margin},k,s}\) and
\(U_{\mathrm{area},k,s}\), and operation-pass ratio
\(r_{\mathrm{stab},k,s}\), all in \([0,1]\):
\begin{align}
S_{\mathrm{force},k,s}
&=1-\lambda_{\mathrm{force}}I_{\mathrm{force},k,s},\\
S_{\mathrm{red},k,s}
&=1-\lambda_{\mathrm{red}}P_{\mathrm{red},k,s},\\
S_{\mathrm{margin},k,s}
&=1-\lambda_{\mathrm{margin}}
+\lambda_{\mathrm{margin}}U_{\mathrm{margin},k,s},\\
S_{\mathrm{area},k,s}
&=1-\lambda_{\mathrm{area}}
+\lambda_{\mathrm{area}}U_{\mathrm{area},k,s},\\
S_{\mathrm{stab},k,s}
&=\left[\epsilon_{\mathrm{stab}}
+(1-\epsilon_{\mathrm{stab}})
r_{\mathrm{stab},k,s}\right]^{\gamma_{\mathrm{stab}}}.
\label{eq:stage_score_components}
\end{align}
The \(\lambda\) terms are component weights, while
\(\epsilon_{\mathrm{stab}}\) and \(\gamma_{\mathrm{stab}}\) control stability
score shaping. The components are combined multiplicatively and gated by hard
feasibility:
\begin{equation}
\begin{aligned}
B_{k,s}&=
S_{\mathrm{force},k,s}
S_{\mathrm{red},k,s}
S_{\mathrm{margin},k,s}
S_{\mathrm{area},k,s}
S_{\mathrm{stab},k,s},\\
Q_{k,s}&=h_{k,s}\left[1-\left(1-B_{k,s}\right)^{\eta_q}\right],
\qquad \eta_q>0.
\end{aligned}
\label{eq:stage_quality}
\end{equation}
Here, \(h_{k,s}=1\) only when mapping, coverage, collision, nominal gravity stability, and all required operation-load checks pass. CoG-uncertainty cases contribute continuously through \(S_{\mathrm{stab},k,s}\). An infeasible candidate receives \(Q_{k,s}=0\).

\subsection{DDPM-Based Initialization}

Many broadly distributed coordinate samples that are feasible in the
canonical state lose feasibility later in the sequence. The DDPM therefore
learns from physics-evaluated configurations to generate support-coordinate
seeds with higher complete-sequence quality for BO initialization. It operates
on the stacked coordinate vector
\(X^{(k)}=[p_{k,1}^{T},p_{k,2}^{T},p_{k,3}^{T}]^{T}\in\mathbb{R}^{9}\), with
the sequence objective \(J_k\) serving as its associated physics-based quality
label. Balloon-hand models are not generated by the DDPM; they are assigned
during physics-based postprocessing.

For a noisy coordinate vector \(X_\ell^{(k)}\) at diffusion step \(\ell\), the
network uses a shared multilayer perceptron with timestep embedding and two
output heads to predict the sampled Gaussian noise
\(\epsilon\in\mathbb{R}^9\) and the sequence objective. The noise-prediction
loss, objective-prediction loss, and total training loss are
\begin{equation}
\begin{aligned}
\mathcal{L}_{\epsilon}
&=\mathbb{E}\!\left[
w_{\epsilon}(J_k)
\left\|\epsilon-\epsilon_{\theta}(X_\ell^{(k)},\ell)\right\|_2^2
\right],\\
\mathcal{L}_{J}
&=\mathbb{E}\!\left[
w_J(J_k)
\left(\hat J_{\theta}(X_\ell^{(k)},\ell)-J_k\right)^2
\right],\\
\mathcal{L}_{\mathrm{DDPM}}
&=\mathcal{L}_{\epsilon}+\lambda_J\mathcal{L}_{J}.
\end{aligned}
\label{eq:ddpm_training_loss}
\end{equation}
Here, \(\theta\) denotes the network parameters, while
\(\epsilon_{\theta}\) and \(\hat J_{\theta}\) are the predicted noise and
sequence objective, respectively. The expectation is over training configurations,
diffusion steps, and sampled noise. The quality-dependent weights
\(w_{\epsilon}(J_k)\) and \(w_J(J_k)\) increase with \(J_k\), assigning greater
training influence to configurations with higher physics-based quality, while
\(\lambda_J\) balances the two losses. In the implementation,
\(w_{\epsilon}(J_k)=0.01+0.99J_k\) and
\(w_J(J_k)=0.2+0.8J_k\).

\begin{figure}[!t]
    \centering
    \includegraphics[width=1\linewidth]{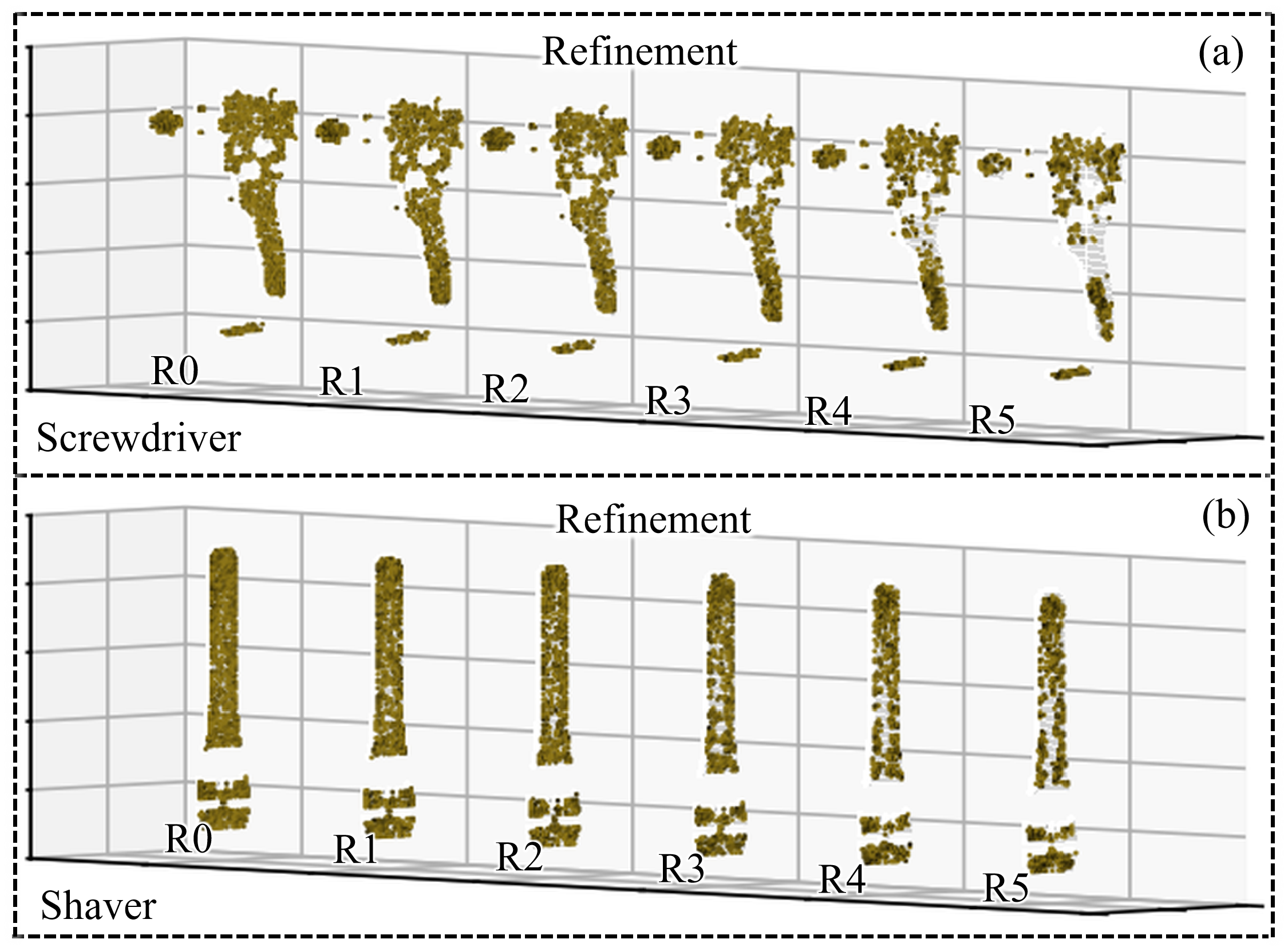}
\caption{DDPM-generated support-coordinate distributions over refinement rounds R0--R5 for (a) the screwdriver and (b) the shaver.}
    \label{fig:Generation_slices}
\end{figure}

At refinement round \(r\), score-guided denoising generates support-coordinate
samples. These samples are mapped to the stage-0 candidate domain and assigned
balloon-hand models to form the complete layouts \(x_k^{(r)}\), which are then
deduplicated and physically evaluated. The 500 unique records with the
highest \(J_k\) are retained as feedback and form the BO seeds
\(\mathcal B_{500}\). Fig.~\ref{fig:Generation_slices} shows how
the generated support-coordinate distribution becomes increasingly
concentrated from R0 to R5 under cross-stage physics feedback.

\subsection{Bayesian Optimization}

The cross-stage objective is treated as an expensive black-box function because it has no readily available analytical gradient and each candidate requires repeated physics-based evaluation. Bayesian optimization addresses this setting by using surrogate models and an acquisition function to select informative candidates with fewer evaluations. Since object geometry, mass properties, operation points, and task loads vary across the predefined stages, one BO run optimizes a fixed canonical support configuration over \(\mathcal{S}_{\mathrm{all}}\). The initial BO seeds are the verified DDPM configurations in \(\mathcal{B}_{500}\). Every subsequent candidate is evaluated at all stages, producing the stage-quality scores \(Q_{k,s}\) for all \(s\in\mathcal{S}_{\mathrm{all}}\), together with the corresponding feasibility results and sequence objective \(J_k\).

\subsubsection{Feature Representation for Bayesian Optimization}

For Gaussian process regression, each candidate is represented by normalized support coordinates, local boundary-margin features, and balloon-hand model indices. 
The BO feature representation of candidate \(x_k\) is defined as
\begin{equation}
\begin{aligned}
\phi(x_k)
=
\big[
&\bar{p}_{k,1}^{T},
\bar{p}_{k,2}^{T},
\bar{p}_{k,3}^{T},
\bar{d}_{k,1},
\bar{d}_{k,2},
\bar{d}_{k,3}, \\
&m_{k,1},
m_{k,2},
m_{k,3}
\big]^{T}
\in \mathbb{R}^{15},
\end{aligned}
\label{eq:bo_feature_vector}
\end{equation}
where \(\bar{p}_{k,i} \in \mathbb{R}^{3}\) denotes the normalized coordinate
of the \(i\)-th canonical suction center of candidate \(x_k\), and
\(\bar{d}_{k,i}\) denotes its normalized boundary-margin feature.
Stage 0 is the fixed canonical reference, keeping the feature dimension and
geometric meaning consistent while \(J_k\) represents complete-sequence
performance. Accordingly,
\(\bar d_{k,i}=d_{\mathrm{edge},k,0,i}/d_{\mathrm{norm}}\), where
\(d_{\mathrm{norm}}\) is the margin-normalization range.

\subsubsection{Geometry--Model Mixed-Kernel Surrogates}

For candidates \(x_k\) and \(x_{k'}\), two GP surrogates sharing the same
kernel are used to model \(Q_{k,0}\) and \(J_k\), respectively. The former
guides canonical physical-quality improvement, while the latter represents
complete-sequence feasibility and robustness.

Because the search space contains both continuous coordinates and discrete
model indices, a mixed kernel is used:
\begin{equation}
\kappa(x_k,x_{k'})
=
\kappa_{\mathrm{cont}}(x_k,x_{k'})
\cdot
\kappa_{\mathrm{model}}(x_k,x_{k'}).
\label{eq:mixed_kernel}
\end{equation}
The corresponding normalized position, normalized margin, and model features
of \(x_k\) and \(x_{k'}\) are denoted by
\(\bar p_{k,i},\bar d_{k,i},m_{k,i}\) and
\(\bar p_{k',i},\bar d_{k',i},m_{k',i}\), respectively. With positive position
and margin length scales \(L_p\) and \(L_d\), the per-hand continuous-feature
distance and Mat\'ern-\(5/2\) kernel are
\begin{equation}
\begin{aligned}
\delta_i(x_k,x_{k'})
&=
\left[
\frac{\|\bar p_{k,i}-\bar p_{k',i}\|_2^2}{L_p^2}
+
\frac{(\bar d_{k,i}-\bar d_{k',i})^2}{L_d^2}
\right]^{1/2},\\
\kappa_{\mathrm{cont}}(x_k,x_{k'})
&=
\prod_{i=1}^{3}
\left(
1+\sqrt{5}\delta_i(x_k,x_{k'})
+\frac{5}{3}\delta_i^2(x_k,x_{k'})
\right)\\
&\quad {}\times
\exp\!\left[
-\sqrt{5}\sum_{i=1}^{3}\delta_i(x_k,x_{k'})
\right].
\end{aligned}
\label{eq:continuous_kernel}
\end{equation}
Let \(\Delta R_{k,k',i}=|R_{\mathrm{suc}}(m_{k,i})-R_{\mathrm{suc}}(m_{k',i})|\). The model kernel uses the fixed scale \(\tau_m=3\) mm:
\begin{equation}
\kappa_{\mathrm{model}}(x_k,x_{k'})
=
\prod_{i=1}^{3}
\exp\left[-\frac{\Delta R_{k,k',i}^2}{2\tau_m^2}\right].
\label{eq:model_kernel}
\end{equation}
Thus, the mixed kernel captures continuous geometric similarity and model-dependent suction-radius differences.

\begin{figure}[!t]
    \centering
    \includegraphics[width=\linewidth]{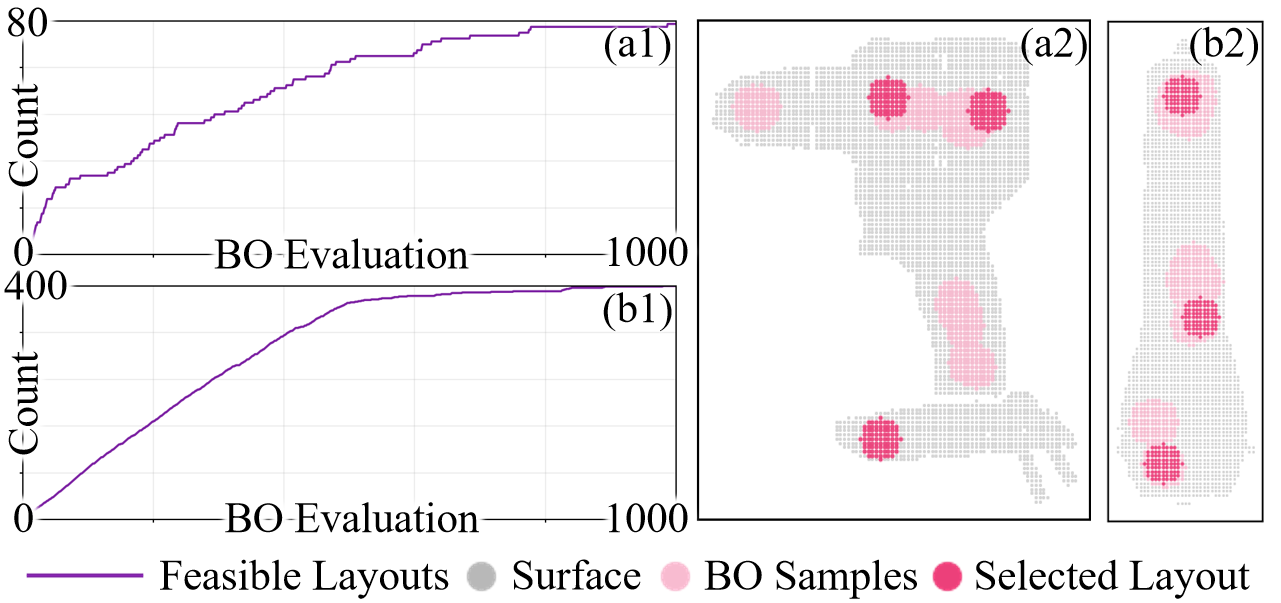}
\caption{BO search results for the screwdriver and shaver. Panels (a1) and
(b1) show the cumulative counts of unique all-stage-feasible configurations; panels
(a2) and (b2) show the explored support locations and selected shared
layouts.}
    \label{fig:boresult}
\end{figure}

\subsubsection{Multi-stage Objective}

To aggregate feasibility across the complete disassembly sequence, an
infeasible candidate \(x_k\) at stage \(s\) is assigned a violation degree
\(v_{k,s}\in[0,1]\). It is the clipped maximum of four stage-level terms: a
binary geometry failure, the normalized shortfall of the stability ratio from
its feasibility threshold, the fraction of failed operation-load cases, and
the complement of the mean normalized boundary margin, weighted by \(1\),
\(0.8\), \(0.8\), and \(0.5\), respectively.
The soft feasibility value of candidate \(x_k\) at stage \(s\) is defined as
\begin{equation}
c_{k,s}
=
\begin{cases}
1, & \text{if } x_k \text{ is feasible at stage } s, \\
\exp\left(-\beta v_{k,s}\right), & \text{otherwise},
\end{cases}
\label{eq:stage_feasibility_score}
\end{equation}
where \(\beta\) controls the penalty strength for infeasible candidates. 

The cross-stage feasibility aggregation is computed from both the average
stage feasibility and the worst-stage feasibility over the complete stage set:
\begin{equation}
C_k
=
w_{\mathrm{mean},C}\frac{1}{|\mathcal{S}_{\mathrm{all}}|}
\sum_{s\in\mathcal{S}_{\mathrm{all}}}c_{k,s}
+
w_{\mathrm{min},C}\min_{s\in\mathcal{S}_{\mathrm{all}}}c_{k,s}.
\label{eq:cross_stage_feasibility}
\end{equation}
Here, \(C_k\) aggregates the feasibility of candidate \(x_k\) across the
complete disassembly sequence. The nonnegative weights
\(w_{\mathrm{mean},C}\) and \(w_{\mathrm{min},C}\), which sum to one, control
the relative effects of average and worst-stage feasibility.

The feasible-stage set is defined below; when it is nonempty, its robust
quality is
\begin{equation}
\begin{aligned}
\mathcal{S}_{\mathrm{feas},k}
&=\{s\in\mathcal{S}_{\mathrm{all}}\mid h_{k,s}=1\},\\
Q_{\mathrm{worst},k}^{\mathrm{feas}}
&=\min_{s\in\mathcal{S}_{\mathrm{feas},k}}Q_{k,s},\\
Q_{\mathrm{mean},k}^{\mathrm{feas}}
&=\frac{1}{|\mathcal{S}_{\mathrm{feas},k}|}
\sum_{s\in\mathcal{S}_{\mathrm{feas},k}}Q_{k,s},\\
Q_{\mathrm{rob},k}&=
w_{\mathrm{worst},Q}
Q_{\mathrm{worst},k}^{\mathrm{feas}}
+
w_{\mathrm{mean},Q}
Q_{\mathrm{mean},k}^{\mathrm{feas}},
\end{aligned}
\label{eq:robust_quality}
\end{equation}
where the nonnegative weights \(w_{\mathrm{worst},Q}\) and
\(w_{\mathrm{mean},Q}\) sum to one. If
\(\mathcal{S}_{\mathrm{feas},k}\) is empty, \(Q_{\mathrm{rob},k}\) is set to
zero.

\begin{figure*}[!t]
    \centering
    \includegraphics[width=1\linewidth]{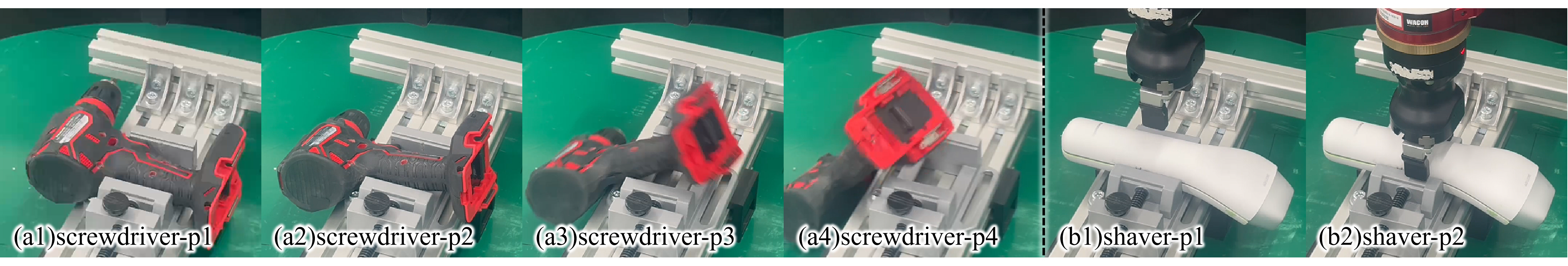}
    \caption{Qualitative evaluation of conventional vise fixturing for the screwdriver and shaver. p1–p4 denote successive observed poses during vise testing.}
    \label{fig:vise}
\end{figure*}

Unlike the single-stage \(Q_{k,s}\), the multi-stage quality \(Q_k\) gates
\(Q_{\mathrm{rob},k}\) by cross-stage feasibility. The final BO objective is
\begin{equation}
\begin{aligned}
Q_k&=C_k^{\gamma_Q}Q_{\mathrm{rob},k},\\
J_k&=w_C C_k+w_Q Q_k,
\qquad w_C+w_Q=1.
\end{aligned}
\label{eq:jk_objective}
\end{equation}
Here, \(\gamma_Q\) is the gate exponent and \(w_C,w_Q\) balance feasibility
and quality.
The parameters in \(J_k\), \(C_k\), and \(Q_k\) are treated as optimization setting parameters for comparative evaluation, rather than fixed physical parameters of the fixturing system.

\subsubsection{Acquisition-Based Candidate Selection}

At iteration \(t\), the surrogates define expected improvements
\(EI_{Q,0,t}\) and \(EI_{J,t}\) for stage-0 quality and the sequence objective.
For the unevaluated candidate set \(\mathcal K_t\), selection is defined by
\begin{equation}
\begin{aligned}
a^{\mathrm{EI}}_{t,k}
&=
\left(1-\alpha_{\mathrm{acq}}\right)
EI_{Q,0,t}(\phi(x_k))\\
&\quad+
\alpha_{\mathrm{acq}}
EI_{J,t}(\phi(x_k)),\\
a_{t,k}&=
a^{\mathrm{EI}}_{t,k}
g_{t,k}^{\mathrm{feas}},\\
k_t^\ast&=\arg\max_{k\in\mathcal{K}_t}a_{t,k}.
\end{aligned}
\label{eq:bo_candidate_selection}
\end{equation}
Here, \(\alpha_{\mathrm{acq}}\in[0,1]\) balances the two objectives, while
\(g_{t,k}^{\mathrm{feas}}\) heuristically favors candidates with higher
estimated physical feasibility. In the implementation,
\(g_{t,k}^{\mathrm{feas}}=0.3+0.7f_{\mathrm{proxy},t,k}\), where
\(f_{\mathrm{proxy},t,k}\in[0,1]\) combines candidate-region validity,
geometric feasibility, and the minimum normalized boundary margin at stage 0.

Equation~\eqref{eq:bo_candidate_selection} is supplemented by a
small-probability exploration step that may select from an acquisition-ranked
subset or the remaining pool. The selected \(x_{k_t^\ast}\) is evaluated at
all stages, and its \(Q_{k_t^\ast,0}\) and \(J_{k_t^\ast}\) update the two
surrogates. After the budget is exhausted, deduplicated archive configurations
are verified across all stages, and the feasible candidate with the highest
\(J_k\) becomes \(x^\ast\). Figs.~\ref{fig:boresult}(a1) and (b1) show the
cumulative unique all-stage-feasible configurations, while (a2) and (b2) show
the explored support locations and selected layouts.

\section{Experiments}
\label{sec:experiments}

The planned layouts are evaluated through three physical experiments: a
qualitative comparison with a conventional vise, robotic screw and component-removal operations, and directional external-load tests. Together, these
experiments examine the practical feasibility, operational loads, and
direction-dependent resistance of the shared fixture layouts. The operational
and directional responses are related in Section~\ref{sec:discussion}.

\subsection{Experimental Setup and Data Processing}

A DynPick six-axis force/torque sensor sampled the operational and externally
applied loads at \SI{50}{Hz}. Each trial was offset-corrected using the initial
stationary signal. For the corrected force \(F=[F_x,F_y,F_z]^T\) and moment
\(M=[M_x,M_y,M_z]^T\), the resultants are
\(|F|=(F_x^2+F_y^2+F_z^2)^{1/2}\) and
\(|M|=(M_x^2+M_y^2+M_z^2)^{1/2}\).
Repeated trials with independent object repositioning were used to account for
variations in placement and contact conditions; the experiment-specific
protocols are given below.

To compare trials with different durations, each detected active-operation
interval was resampled to 0--100\% operation progress. The resultant force and
moment histories were then smoothed using a \SI{0.1}{s} moving window before
the across-trial mean and standard deviation were calculated.

Before each screw-operation or component-removal trial, the robotic end
effector was manually aligned with the target screw or component. The robot
then executed the prescribed screw-removal, tightening, or component-removal
operation.

\subsection{Fixture Comparison, Disassembly Feasibility, and Operational Loads}

Fig.~\ref{fig:vise} illustrates the limitations of a parallel-jaw vise under the pose requirement that the screws remain vertical and accessible.
In Figs.~\ref{fig:vise}(a1), (a2), (a3), and (a4), the curved screwdriver
housing forms only narrow line or limited local surface contact with the jaws.
Combined with its asymmetric geometry and mass distribution, this contact
causes progressive tilting and eventual loss of fixation. In
Figs.~\ref{fig:vise}(b1) and (b2), the elongated shaver is held securely, but
the jaws obstruct the removable upper cover and leave insufficient gripper
clearance. Thus, a rigid vise may either insufficiently constrain an irregular
product or obstruct the workspace required for disassembly.

\begin{figure*}[!t]
    \centering
    \includegraphics[width=\linewidth]{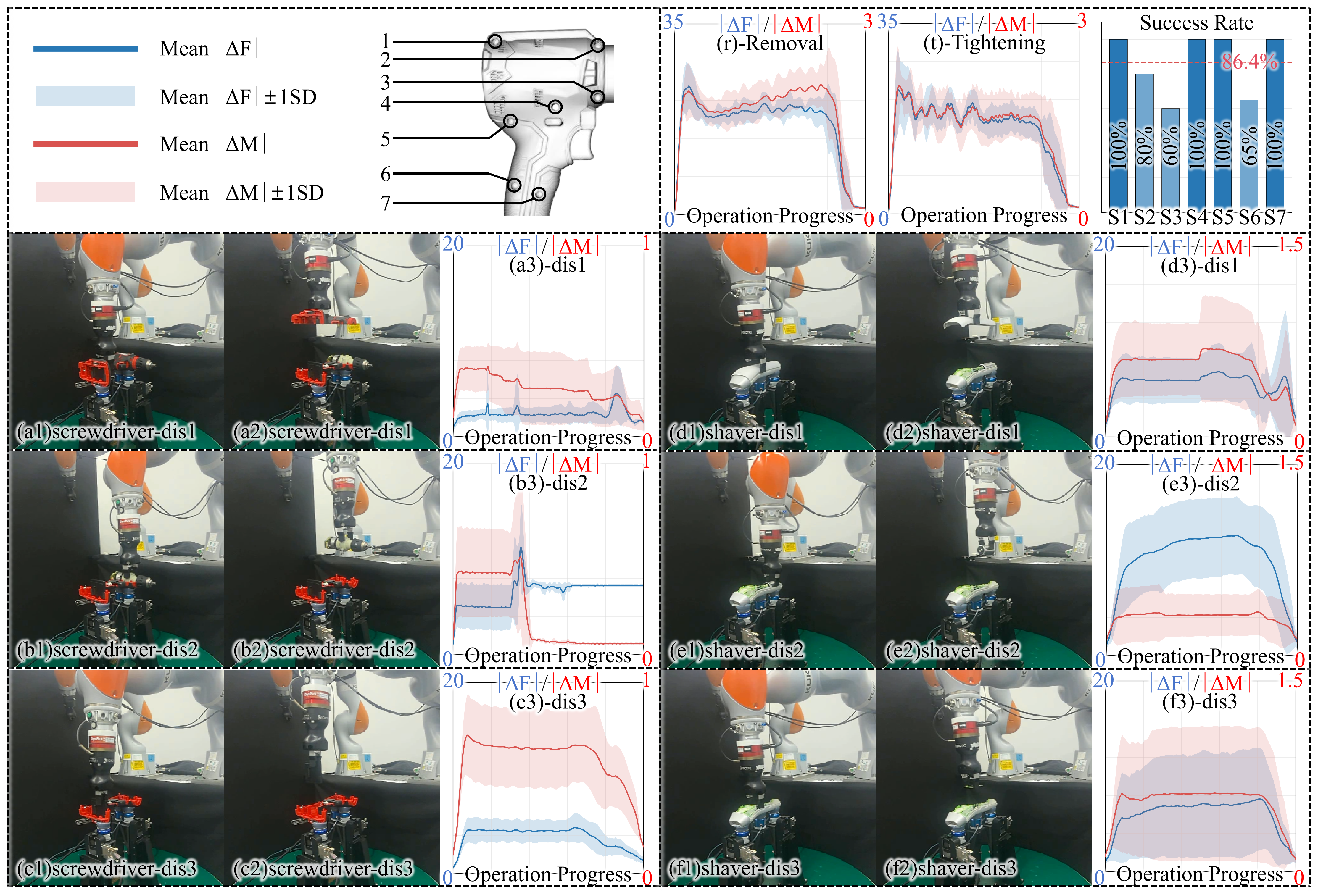}
    \caption{Screw-operation and component-removal experiments. The plots show the mean resultant force and moment with one-standard-deviation bands; the bars report screw-removal success rates.}
    \label{fig:exp2}
\end{figure*}

\begin{table*}[!t]
\centering
\caption{Mean six-axis operational peak responses during screw operations and component removal.}
\label{tab:operational_wrenches}
\normalsize
\setlength{\tabcolsep}{3pt}
\begin{tabular*}{\textwidth}{@{\extracolsep{\fill}} c c c c c c c c c c}
\toprule
& & \multicolumn{5}{c}{Screwdriver} & \multicolumn{3}{c}{Shaver} \\
\cmidrule(lr){3-7}\cmidrule(lr){8-10}
Response & Unit & Removal & Tightening & dis-1 & dis-2 & dis-3 & dis-1 & dis-2 & dis-3 \\
\midrule
$|F_x|$ & N & 23.44 & 23.59 & 1.17 & 1.39 & 3.47 & 3.34 & 1.81 & 2.49 \\
$|F_y|$ & N & 1.55 & 1.66 & 1.71 & 3.34 & 2.17 & 3.76 & 1.72 & 1.67 \\
$|F_z|$ & N & 11.80 & 13.62 & 7.42 & 13.44 & 3.13 & 9.11 & 12.44 & 7.60 \\
$|F|$ & N & \cellcolor{ForceHighlight}26.24 & \cellcolor{ForceHighlight}27.20 & \cellcolor{ForceHighlight}7.77 & \cellcolor{ForceHighlight}13.96 & \cellcolor{ForceHighlight}5.06 & \cellcolor{ForceHighlight}10.07 & \cellcolor{ForceHighlight}12.60 & \cellcolor{ForceHighlight}8.05 \\
\cmidrule(lr){1-10}
$|M_x|$ & N$\cdot$m & 0.31 & 0.30 & 0.24 & 0.55 & 0.35 & 0.50 & 0.23 & 0.38 \\
$|M_y|$ & N$\cdot$m & 2.29 & 2.20 & 0.24 & 0.37 & 0.58 & 0.47 & 0.23 & 0.53 \\
$|M_z|$ & N$\cdot$m & 0.43 & 0.54 & 0.30 & 0.06 & 0.31 & 0.31 & 0.15 & 0.16 \\
$|M|$ & N$\cdot$m & \cellcolor{MomentHighlight}2.33 & \cellcolor{MomentHighlight}2.30 & \cellcolor{MomentHighlight}0.41 & \cellcolor{MomentHighlight}0.66 & \cellcolor{MomentHighlight}0.72 & \cellcolor{MomentHighlight}0.74 & \cellcolor{MomentHighlight}0.32 & \cellcolor{MomentHighlight}0.68 \\
\bottomrule
\end{tabular*}
\end{table*}

The planned shared fixture layout was next evaluated through screw removal and
tightening at the seven
locations shown in Fig.~\ref{fig:exp2}. An axial force of
\SI{18}{N} was applied in the negative \(z\)-direction during removal, whereas
a tightening torque of \SI{0.5}{N.m} was applied during tightening.
Figs.~\ref{fig:exp2}(r) and (t) show the mean resultant force and moment
histories across the seven locations for removal and tightening, respectively;
the shaded regions represent one standard deviation among locations. The mean
peak resultant force and moment were \SI{26.24}{N} and \SI{2.33}{N.m} during
removal and \SI{27.20}{N} and \SI{2.30}{N.m} during tightening, respectively.
In these load-measurement trials, removal and tightening were completed at all
seven locations without changing the shared fixture layout, and the recorded
load histories exhibited consistent process trends.

To evaluate repeatability, 20 screw-removal attempts were conducted at each
location. The success rates at locations 1--7 were 100\%, 80\%, 60\%, 100\%,
100\%, 65\%, and 100\%, respectively. Overall, 121 of 140 attempts succeeded;
in Fig.~\ref{fig:exp2}, the bars report the location-wise rates, while the
dashed line indicates the aggregate rate of 86.4\%. The observed failures were primarily
associated with misalignment between the screwdriver bit and the screw.
Because target detection and autonomous alignment are outside the scope of
this study, the reported rate characterizes the complete screw-removal
operation; the alignment failures are not attributed to the fixture layout.

Sequential removal of three components was evaluated for each object using 10
independent repositioning trials per stage. The highest and lowest response
peaks were excluded, and the remaining eight were averaged. For the
screwdriver, Figs.~\ref{fig:exp2}(a1), (a2), (b1), (b2), (c1), and (c2)
show the pre- and post-removal states, while (a3), (b3), and (c3) show the
corresponding load histories. The respective shaver results are shown in
Figs.~\ref{fig:exp2}(d1), (d2), (e1), (e2), (f1), and (f2), with the load
histories in (d3), (e3), and (f3). The histories are normalized by operation
progress, with shading indicating one standard deviation. All 60 trials
(30 per object) successfully removed the prescribed components without
changing the planned shared fixture layout, demonstrating its feasibility
throughout the sequential disassembly stages.

\begin{figure*}[!t]
    \centering
    \includegraphics[width=1\linewidth]{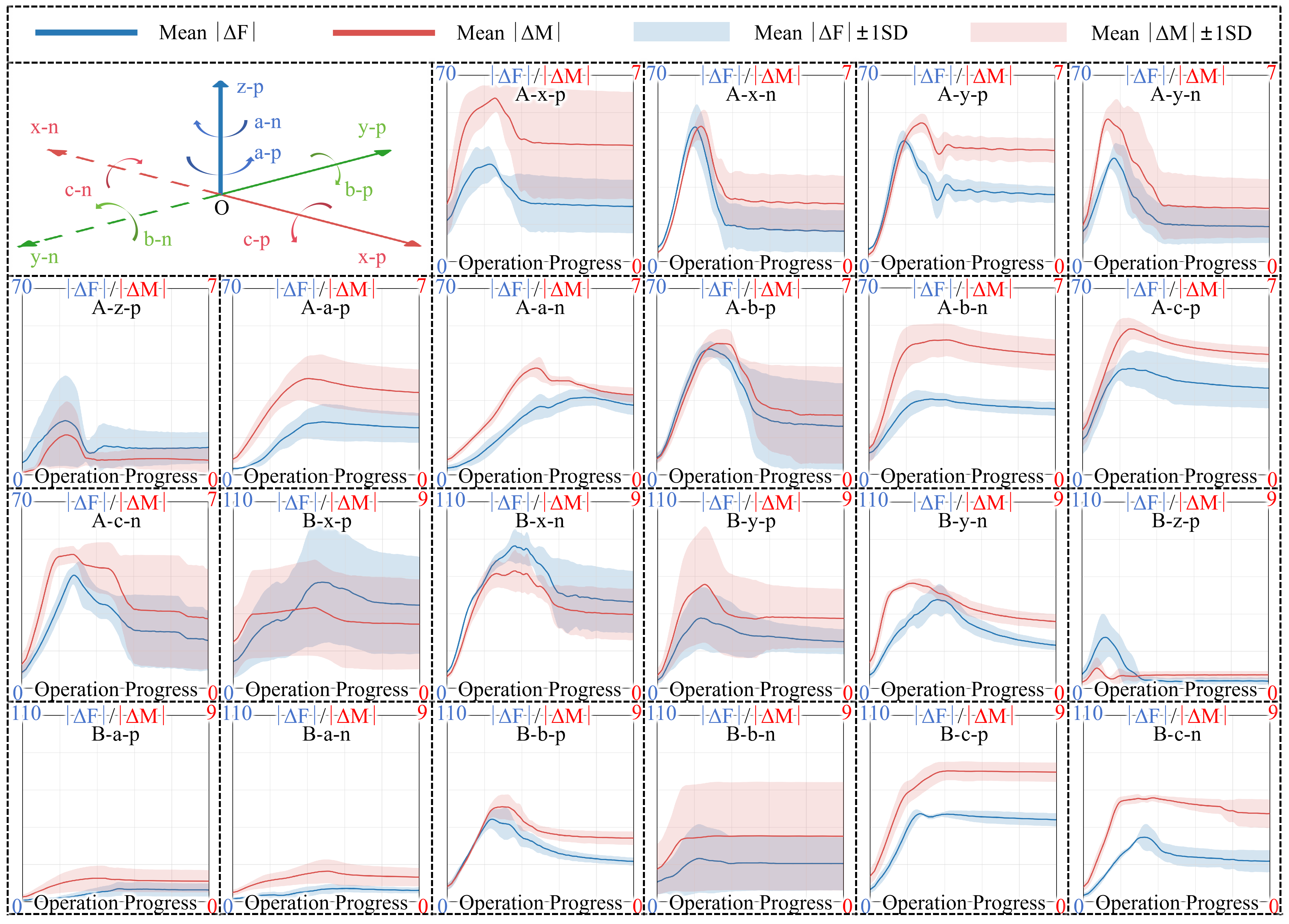}
    \caption{Directional-load responses in the 11 tests for (A) the screwdriver and (B) the shaver. Curves and bands show the mean incremental resultant loads and one standard deviation.}
    \label{fig:exp33}
\end{figure*}

\begin{table*}[!t]
\centering
\caption{Mean peak responses in the 11 directional-load tests, reported as screwdriver/shaver.}
\label{tab:directional_limits}
\normalsize
\setlength{\tabcolsep}{2.0pt}
\begin{tabular*}{\textwidth}{@{\extracolsep{\fill}} c c c c c c c c c}
\toprule
\multirow{2}{*}{Direction} &
\multicolumn{4}{c}{Peak force response (N)} &
\multicolumn{4}{c}{Peak moment response (N$\cdot$m)} \\
\cmidrule(lr){2-5}\cmidrule(lr){6-9}
& $|\Delta F_x|$ & $|\Delta F_y|$ & $|\Delta F_z|$ & $|\Delta F|$
& $|\Delta M_x|$ & $|\Delta M_y|$ & $|\Delta M_z|$ & $|\Delta M|$ \\
\midrule
$x\text{-p}$ & 6.07/14.54 & 36.68/45.36 & 13.01/70.53 & \cellcolor{ForceHighlight}37.20/79.09 & 5.04/5.01 & 1.48/0.79 & 2.16/2.19 & 5.61/5.47 \\
$x\text{-n}$ & 7.65/18.52 & 27.38/24.72 & 48.80/84.86 & \cellcolor{ForceHighlight}56.13/88.75 & 4.62/4.86 & 2.72/4.16 & 1.45/0.35 & 5.46/6.03 \\
$y\text{-p}$ & 28.35/36.99 & 9.44/15.79 & 42.04/42.53 & \cellcolor{ForceHighlight}50.79/52.97 & 2.18/3.71 & 4.96/5.00 & 0.85/0.72 & 5.29/6.26 \\
$y\text{-n}$ & 31.12/31.94 & 15.05/8.45 & 16.09/53.55 & \cellcolor{ForceHighlight}38.64/58.02 & 1.55/2.42 & 4.86/4.96 & 0.90/0.40 & 5.12/5.45 \\
$z\text{-p}$ & 9.65/6.52 & 7.59/4.45 & 25.33/32.17 & \cellcolor{ForceHighlight}27.62/32.49 & 1.02/0.47 & 1.74/0.83 & 0.39/0.44 & 2.02/1.02 \\
\midrule
$a\text{-p}$ & 3.87/2.58 & 9.48/3.19 & 18.28/7.35 & 20.70/7.82 & 2.01/0.34 & 1.68/0.43 & 2.77/1.48 & \cellcolor{MomentHighlight}3.79/1.51 \\
$a\text{-n}$ & 12.94/4.77 & 7.42/4.14 & 27.79/10.61 & 29.95/11.45 & 1.76/0.65 & 2.94/0.87 & 3.10/1.92 & \cellcolor{MomentHighlight}4.55/2.01 \\
$b\text{-p}$ & 6.79/6.96 & 25.11/25.72 & 45.46/47.53 & 52.01/54.18 & 4.67/4.42 & 2.81/2.12 & 1.39/0.95 & \cellcolor{MomentHighlight}5.60/4.89 \\
$b\text{-n}$ & 3.95/7.75 & 29.83/41.25 & 8.04/16.11 & 30.19/44.54 & 5.08/5.26 & 0.83/1.17 & 0.87/0.71 & \cellcolor{MomentHighlight}5.18/5.44 \\
$c\text{-p}$ & 26.97/36.09 & 10.04/19.00 & 37.30/44.13 & 46.05/57.32 & 1.38/3.94 & 4.86/5.00 & 0.63/0.35 & \cellcolor{MomentHighlight}4.99/6.38 \\
$c\text{-n}$ & 35.09/34.51 & 11.99/4.29 & 20.76/20.30 & 42.00/38.61 & 0.78/1.74 & 5.01/4.96 & 0.98/0.68 & \cellcolor{MomentHighlight}5.11/5.26 \\
\bottomrule
\end{tabular*}
\end{table*}

Table~\ref{tab:operational_wrenches} summarizes the mean operational
peak responses for all tasks. The resultant-force and resultant-moment entries
are shaded in light blue and light red, respectively. For the screwdriver, the mean peak resultant
forces in dis-1, dis-2, and dis-3 were \SI{7.77}{N}, \SI{13.96}{N}, and
\SI{5.06}{N}, respectively, while the corresponding mean peak resultant
moments were \SI{0.41}{N.m}, \SI{0.66}{N.m}, and \SI{0.72}{N.m}. For the
shaver, the mean peak resultant forces were \SI{10.07}{N}, \SI{12.60}{N},
and \SI{8.05}{N}, while the corresponding mean peak resultant moments were
\SI{0.74}{N.m}, \SI{0.32}{N.m}, and \SI{0.68}{N.m}. The largest
component-removal force occurred at dis-2 for both objects, whereas the largest
component-removal moment occurred at dis-3 for the screwdriver and dis-1 for
the shaver. Thus, the largest force and moment responses did not necessarily
occur at the same disassembly stage.

The table also retains the three measured force and moment components. Screw
loads persist over the active operation, whereas component-removal loads are
concentrated between gripper contact and component separation. Differences
among objects and stages further reflect changes in geometry, support pose,
removed component, and contact location. The following experiment complements
these task demands by measuring direction-dependent load resistance.

\subsection{Directional External-Load Resistance}

Directional resistance was measured in five translational directions
(\(x\text{-p}\), \(x\text{-n}\), \(y\text{-p}\), \(y\text{-n}\), and
\(z\text{-p}\)) and six rotational directions
(\(a\text{-p}\), \(a\text{-n}\), \(b\text{-p}\), \(b\text{-n}\),
\(c\text{-p}\), and \(c\text{-n}\)). Each direction was tested in five trials
with independent object repositioning. The upper-left schematic in
Fig.~\ref{fig:exp33} defines these directions. Here,
\(x,y,z\) and \(a,b,c\) denote the commanded translational and rotational
motions of the robot end effector, respectively, rather than individual sensor
axes. The suffixes p and n denote positive and negative commands, and A and B
identify the screwdriver and shaver, respectively.

Because the fixture supports the object from below, loading in the
\(z\text{-n}\) direction further compresses the balloon hands and their linear
modules rather than inducing motion away from or out of the support region.
Increasing this load could damage the linear modules without producing a
meaningful loss-of-support condition. Therefore, \(z\text{-n}\) was excluded,
whereas \(z\text{-p}\) characterizes resistance to upward separation.

The median six-axis wrench immediately before loading was subtracted to remove
fixture preload and static object weight.
Let \(\Delta F_j\) and \(\Delta M_j\), \(j\in\{x,y,z\}\), denote the resulting
incremental force and moment components, respectively. The corresponding
incremental resultants \(|\Delta F|\) and \(|\Delta M|\) use the same Euclidean
norms as the operational loads.
In each trial, the robot end effector moved at a fixed rate in the commanded
direction until an observable loss of stable support occurred. For the
\(z\text{-p}\) tests, loading was instead terminated when the gripper detached
from the object, while the object remained securely retained by the balloon
hands. The gripper attachment region was kept similar across trials. After
progress normalization and \SI{0.1}{s} moving-window smoothing, the
pre-termination peaks of all incremental components and resultants were
extracted.

The response panels in Fig.~\ref{fig:exp33} show the mean incremental
resultant force and moment histories in the 11 loading directions, with the
shaded regions representing one standard deviation over the five trials.
Table~\ref{tab:directional_limits} further summarizes the mean peak
incremental force components, moment components, and their resultants. Because
the commanded motion need not coincide with a sensor axis and its application
point is offset from the sensor, the tests produce coupled force--moment
responses. The component columns preserve this measured coupling rather than
representing independent limits. Light-blue \(|\Delta F|\) entries and
light-red \(|\Delta M|\) entries denote the primary translational and
rotational resistance responses, respectively.

\begin{table*}[!t]
\centering
\caption{Direction-wise empirical stability margins \(\xi_{d,o}\) (\%) under the unfavorable-direction assumption, reported as screwdriver/shaver.}
\label{tab:operational_load_margin}
\footnotesize
\setlength{\tabcolsep}{1.8pt}
\renewcommand{\arraystretch}{1.18}
\newcommand{\marginpair}[2]{#1\,/\,#2}
\begin{tabular*}{\textwidth}{@{\extracolsep{\fill}} c c c c c c c c c c c c c}
\toprule
\multirow{2}{*}{Operation} &
\multicolumn{5}{c}{Translational direction} &
\multicolumn{6}{c}{Rotational direction} &
\multirow{2}{*}{Mean} \\
\cmidrule(lr){2-6}\cmidrule(lr){7-12}
& \(x\text{-p}\) & \(x\text{-n}\) & \(y\text{-p}\) & \(y\text{-n}\) & \(z\text{-p}\)
& \(a\text{-p}\) & \(a\text{-n}\) & \(b\text{-p}\) & \(b\text{-n}\) & \(c\text{-p}\) & \(c\text{-n}\) & \\
\midrule
Removal
& \marginpair{29.5}{--} & \marginpair{53.3}{--} & \marginpair{48.3}{--} & \marginpair{32.1}{--} & \marginpair{\worstval{5.0}}{--}
& \marginpair{38.5}{--} & \marginpair{48.8}{--} & \marginpair{58.4}{--} & \marginpair{55.0}{--} & \marginpair{53.3}{--} & \marginpair{54.4}{--} & \marginpair{43.3}{--} \\
Tightening
& \marginpair{\worstval{26.9}}{--} & \marginpair{51.5}{--} & \marginpair{46.4}{--} & \marginpair{29.6}{--} & \marginpair{\worstval{1.5}}{--}
& \marginpair{39.3}{--} & \marginpair{49.5}{--} & \marginpair{58.9}{--} & \marginpair{55.6}{--} & \marginpair{53.9}{--} & \marginpair{55.0}{--} & \marginpair{42.6}{--} \\
dis-1
& \marginpair{79.1}{87.3} & \marginpair{86.2}{88.7} & \marginpair{84.7}{81.0} & \marginpair{79.9}{82.6} & \marginpair{71.9}{69.0}
& \marginpair{89.2}{\worstval{51.0}} & \marginpair{91.0}{63.2} & \marginpair{\bestval{92.7}}{84.9} & \marginpair{\bestval{92.1}}{86.4} & \marginpair{91.8}{88.4} & \marginpair{\bestval{92.0}}{85.9} & \marginpair{86.4}{78.9} \\
dis-2
& \marginpair{62.5}{84.1} & \marginpair{75.1}{85.8} & \marginpair{72.5}{76.2} & \marginpair{63.9}{78.3} & \marginpair{49.5}{\worstval{61.2}}
& \marginpair{82.6}{78.8} & \marginpair{85.5}{84.1} & \marginpair{88.2}{93.5} & \marginpair{87.3}{\bestval{94.1}} & \marginpair{86.8}{\bestval{95.0}} & \marginpair{87.1}{\bestval{93.9}} & \marginpair{76.4}{84.1} \\
dis-3
& \marginpair{86.4}{89.8} & \marginpair{91.0}{90.9} & \marginpair{90.0}{84.8} & \marginpair{86.9}{86.1} & \marginpair{81.7}{75.2}
& \marginpair{81.0}{\worstval{55.0}} & \marginpair{84.2}{66.2} & \marginpair{87.1}{86.1} & \marginpair{86.1}{87.5} & \marginpair{85.6}{89.3} & \marginpair{85.9}{87.1} & \marginpair{86.0}{81.6} \\
\bottomrule
\end{tabular*}
\end{table*}

For the screwdriver, the translational and rotational resultants span
\SIrange{27.62}{56.13}{N} and \SIrange{3.79}{5.60}{N.m}, respectively; the
corresponding shaver ranges are \SIrange{32.49}{88.75}{N} and
\SIrange{1.51}{6.38}{N.m}.

In both \(z\text{-p}\) tests, the gripper detached while the balloon hands
retained the object. Thus, \SI{27.62}{N} and \SI{32.49}{N} are the peak loads
reached by the present loading interface, not upward failure limits; the
fixtures sustained at least these measured loads.

Both objects showed relatively large translational resistance in
\(x\text{-n}\), but different rotational anisotropy. The screwdriver's
L-shaped geometry and offset mass distribution produce direction-dependent
moment arms. The elongated shaver instead has markedly different support spans
and resisting moment arms, yielding lower responses about \(a\) than about
\(b\) or \(c\). Although baseline correction removes static weight from the
signals, mass still affects balloon-hand compression, contact area, and normal
preload. The observed anisotropy therefore reflects object geometry, mass
distribution, CoG location, support configuration, contact conditions, and
load-application location.

These measurements provide the directional resistance references used in the
subsequent operational-load analysis.

\section{Computational Evaluation and Discussion}
\label{sec:discussion}

This section relates operational loads to directional responses to quantify
empirical stability margins, compares the multi-stage objective settings, and
evaluates DDPM-based BO initialization through a controlled ablation.

\subsection{Operational Loads and Empirical Stability Margins}

The direction-dependent responses measured in Section~\ref{sec:experiments}
serve as empirical resistance references for the tested layouts and loading
conditions. They are not interpreted as universal fixture-failure limits.

Table~\ref{tab:operational_wrenches} reports resultants from coupled six-axis
task loads, whereas the directional tests use predefined translational and
rotational motions. Because an operational wrench does not correspond uniquely
to one test direction, an unfavorable-loading assumption compares its complete
mean force resultant separately with all five translational responses and its
complete mean moment resultant with all six rotational responses. This does
not imply simultaneous loading in every direction; it tests each directional
response against the complete operational demand, whose projection onto any
direction cannot exceed its resultant. For operation \(o\), let \(D_o\) denote
\(|F|_o\) for a translational comparison or \(|M|_o\) for a rotational
comparison, and let \(R_d\) denote the corresponding mean resultant response
measured in direction \(d\). The direction-wise empirical stability margin is
defined as
\begin{equation}
\xi_{d,o}=100\left(1-\frac{D_o}{R_d}\right).
\label{eq:empirical_operational_margin}
\end{equation}
A larger \(\xi_{d,o}\) indicates a larger measured directional resistance
margin relative to the operational load.

The operation-level mean empirical stability margin is defined as the
equal-weight average over the 11 measured directions:
\begin{equation}
\bar\xi_o=\frac{1}{11}\sum_{d\in\mathcal D}\xi_{d,o}.
\label{eq:mean_empirical_margin}
\end{equation}
No task-specific probability or importance distribution is assumed for the
external disturbance directions; therefore, each measured direction is
assigned equal weight. Because \(\xi_{d,o}\) is a dimensionless normalized
margin, the translational and rotational results can be aggregated without
mixing their original force and moment units. The resulting mean provides a
descriptive operation-level indicator, while the individual direction-wise
values are retained to show anisotropic behavior.
The object-level mean empirical stability margin is the mean of \(\bar\xi_o\)
over all tested operations for that object. The direction-wise and mean margins
quantify the measured stability reserve of each shared fixture layout under the
tested disassembly operations and loading conditions.

\begingroup
\colorlet{BestShade}{white}
\colorlet{WorstShade}{white}

\begin{table*}[!t]
\centering
\caption{Multi-stage BO objective settings and ablation results.}
\label{tab:jk_preset_results}
\small
\setlength{\tabcolsep}{2.0pt}
\renewcommand{\arraystretch}{1.00}
\begin{tabular*}{\textwidth}{@{\extracolsep{\fill}}cc@{\hspace{3pt}}|cccc@{\hspace{3pt}}|ccc@{\hspace{3pt}}|cc@{\hspace{3pt}}|c}
\toprule
\multirow{2}{*}{Combination} &
\multirow{2}{*}{Statistic} &
\multicolumn{4}{c|}{Stage quality} &
\multicolumn{3}{c|}{Cross-stage behavior} &
\multicolumn{2}{c|}{Objective distribution} &
\multicolumn{1}{c}{Best new support} \\
\cmidrule(lr){3-6}
\cmidrule(lr){7-9}
\cmidrule(lr){10-11}
\cmidrule(lr){12-12}
& & \(Q_0\) & \(Q_1\) & \(Q_2\) & \(Q_3\) &
\(r_{4/4}\) (\%) &
\(\bar r_{\mathrm{feas}}^{\,10\%}\) &
\(\bar v^{\,10\%}\) &
\(\bar J_k^{\,10\%}\) &
\(\bar J_k^{\mathrm{top100}}\) &
\(Q_{\mathrm{mean}}^{\mathrm{new},\ast}\) \\
\midrule
\multirow{2}{*}{\shortstack{J1\\(0.60,0.70,0.60)}} & Trimmed mean & 0.977 & 0.982 & 0.183 & 0.548 & 51.5 & 0.795 & 0.123 & 0.604 & 0.820 & -- \\
 & Best new & 0.966 & 0.973 & 0.567 & 0.939 & -- & -- & -- & -- & -- & 0.861 \\
\cmidrule(lr){1-12}
\multirow{2}{*}{\shortstack{J2\\(0.50,0.70,0.60)}} & Trimmed mean & 0.979 & 0.984 & \weakval{0.148} & \weakval{0.471} & \weakval{43.2} & \worstval{0.751} & \worstval{0.149} & \worstval{0.529} & \worstval{0.775} & -- \\
 & Best new & 0.967 & 0.973 & 0.464 & 0.940 & -- & -- & -- & -- & -- & 0.836 \\
\cmidrule(lr){1-12}
\multirow{2}{*}{\shortstack{J3\\(0.70,0.70,0.60)}} & Trimmed mean & \worstval{0.975} & \worstval{0.981} & 0.194 & 0.598 & 55.4 & 0.818 & 0.110 & 0.660 & 0.865 & -- \\
 & Best new & \bestval{0.985} & \bestval{0.988} & \keyval{0.978} & \keyval{0.978} & -- & -- & -- & -- & -- & \keyval{0.982} \\
\cmidrule(lr){1-12}
\multirow{2}{*}{\shortstack{J4\\(0.80,0.70,0.60)}} & Trimmed mean & 0.975 & 0.981 & 0.203 & 0.624 & \keyval{57.7} & 0.831 & 0.100 & \keyval{0.711} & 0.911 & -- \\
 & Best new & 0.980 & 0.985 & \keyval{0.980} & \keyval{0.976} & -- & -- & -- & -- & -- & \keyval{0.980} \\
\cmidrule(lr){1-12}
\multirow{2}{*}{\shortstack{J5\\(0.60,0.50,0.60)}} & Trimmed mean & \worstval{0.975} & \worstval{0.981} & 0.200 & 0.608 & 56.8 & 0.829 & 0.101 & 0.593 & 0.819 & -- \\
 & Best new & \worstval{0.961} & \worstval{0.970} & \worstval{0.456} & \worstval{0.919} & -- & -- & -- & -- & -- & \worstval{0.827} \\
\cmidrule(lr){1-12}
\multirow{2}{*}{\shortstack{J6\\(0.60,0.60,0.60)}} & Trimmed mean & 0.976 & 0.982 & 0.197 & 0.584 & 54.9 & 0.815 & 0.111 & 0.601 & 0.820 & -- \\
 & Best new & 0.966 & 0.973 & 0.567 & 0.939 & -- & -- & -- & -- & -- & 0.861 \\
\cmidrule(lr){1-12}
\multirow{2}{*}{\shortstack{J7\\(0.60,0.80,0.60)}} & Trimmed mean & 0.979 & 0.984 & \weakval{0.153} & \weakval{0.479} & \weakval{44.7} & 0.757 & 0.147 & 0.595 & 0.820 & -- \\
 & Best new & 0.966 & 0.973 & 0.567 & 0.939 & -- & -- & -- & -- & -- & 0.861 \\
\cmidrule(lr){1-12}
\multirow{2}{*}{\shortstack{J8\\(0.60,0.70,0.50)}} & Trimmed mean & 0.976 & 0.982 & 0.187 & 0.575 & 53.1 & 0.804 & 0.118 & 0.624 & 0.838 & -- \\
 & Best new & \bestval{0.985} & \bestval{0.988} & \keyval{0.978} & \keyval{0.978} & -- & -- & -- & -- & -- & \keyval{0.982} \\
\cmidrule(lr){1-12}
\multirow{2}{*}{\shortstack{J9\\(0.60,0.70,0.70)}} & Trimmed mean & 0.978 & 0.983 & \weakval{0.167} & \weakval{0.509} & 47.7 & 0.775 & 0.136 & 0.574 & 0.802 & -- \\
 & Best new & 0.966 & 0.973 & 0.567 & 0.939 & -- & -- & -- & -- & -- & 0.861 \\
\cmidrule(lr){1-12}
\multirow{2}{*}{\shortstack{J10\\(0.60,0.70,0.80)}} & Trimmed mean & \bestval{0.980} & \bestval{0.985} & \keyval{0.273} & \keyval{0.726} & \weakval{46.0} & \bestval{0.863} & \bestval{0.069} & 0.644 & \bestval{0.978} & -- \\
 & Best new & 0.968 & 0.974 & 0.944 & 0.944 & -- & -- & -- & -- & -- & 0.957 \\
\bottomrule
\end{tabular*}
\end{table*}

\begin{table*}[!t]
\centering
\caption{DDPM-seed ablation under the fixed J4 BO setting.}
\label{tab:dp_seed_ablation}
\small
\setlength{\tabcolsep}{2.2pt}
\begin{tabular*}{\textwidth}{@{\extracolsep{\fill}}l@{\hspace{2pt}}|cccc@{\hspace{2pt}}|rrrrr}
\toprule
Initialization &
\multicolumn{4}{c|}{Trimmed-mean Stage quality} &
\(r_{4/4}\) (\%) &
\(\bar r_{\mathrm{feas}}^{\,10\%}\) &
\(\bar v^{\,10\%}\) &
\(\bar J_k^{\,10\%}\) &
\(N_{4/4}^{\mathrm{uniq}}\) \\
\cmidrule(lr){2-5}
& \(Q_0\) & \(Q_1\) & \(Q_2\) & \(Q_3\) & & & & & \\
\midrule
Without DDPM seeds & \cellcolor{WorstShade}0.965 & \cellcolor{WorstShade}0.972 & \cellcolor{WorstShade}0.000 & \cellcolor{WorstShade}0.000 & \cellcolor{WorstShade}0.0 & \cellcolor{WorstShade}0.500 & \cellcolor{WorstShade}0.386 & \cellcolor{WorstShade}0.348 & \cellcolor{WorstShade}0 \\
With DDPM seeds    & \cellcolor{BestShade}0.975 & \cellcolor{BestShade}0.981 & \cellcolor{BestShade}0.203 & \cellcolor{BestShade}0.624 & \cellcolor{BestShade}57.7 & \cellcolor{BestShade}0.831 & \cellcolor{BestShade}0.100 & \cellcolor{BestShade}0.711 & \cellcolor{BestShade}79 \\
\bottomrule
\end{tabular*}
\end{table*}

\endgroup

Table~\ref{tab:operational_load_margin} reports the resulting margins.
For the screwdriver, screw removal and tightening have mean force margins of
33.6\% and 31.2\% across the five translational directions and mean moment
margins of 51.4\% and 52.0\% across the six rotational directions. The
corresponding operation-level mean empirical stability margins are 43.3\% and
42.6\%.

The operation-level mean empirical stability margins for screwdriver dis-1,
dis-2, and dis-3 are 86.4\%, 76.4\%, and 86.0\%, respectively. The
corresponding shaver margins are 78.9\%, 84.1\%, and 81.6\%. Some of the
comparatively smaller shaver
margins in \(a\text{-p}\) arise from its lower measured rotational response
in that direction, consistent with the anisotropy shown in
Fig.~\ref{fig:exp33}.

Averaging the operation-level values gives mean empirical stability margins of
66.9\% for the screwdriver and 81.6\% for the shaver. Green entries indicate
comparatively larger direction-wise margins, whereas yellow entries indicate
comparatively smaller margins for each object. These results depend on the tested shared fixture layouts, object
geometry and mass distribution, contact states, load-application locations,
and unfavorable-loading assumption. All direction-wise margins remain
positive, indicating that the measured directional responses exceed the
corresponding operational demands under the defined assumption.

\subsection{Multi-Stage Bayesian Optimization Ablation}
\label{sec:bo_evaluation}

To select the objective setting for multi-stage BO, ten weight combinations,
denoted J1--J10, were evaluated on the screwdriver because its asymmetric
geometry and disassembly sequence produce substantial cross-stage changes in
the candidate region, mass, and CoG. Only the weights of \(C_k\),
\(Q_{\mathrm{rob},k}\), and \(J_k\) in
Eqs.~\eqref{eq:cross_stage_feasibility}--\eqref{eq:jk_objective} were varied.
All settings used the same 500 DDPM seeds, BO random seed, and 1000 subsequent
evaluations over all four stages. In Table~\ref{tab:jk_preset_results}, each
triplet gives \((w_C,w_{\mathrm{mean},C},w_{\mathrm{worst},Q})\); the reported
statistics exclude the common initialization seeds, and the trimmed means
remove the lowest and highest 10\%; the top-100 mean averages the 100 highest
\(J_k\) values.

The later-stage qualities distinguish the settings because component removal
substantially changes the available support region and mass properties. J10
achieves the highest stage-2 and stage-3 trimmed-mean qualities but only a
46.0\% all-stage-feasible rate. In contrast, J4 achieves the highest
all-stage-feasible rate of 57.7\%, with a mean feasible-stage ratio of 0.831,
a mean violation degree of 0.100, and a best-new-support mean quality of 0.980.
J4 is therefore selected because it concentrates the search in regions that
remain feasible throughout the sequence while retaining high support quality.

\subsection{DDPM-Initialization Ablation}
\label{sec:selected_setting_dp_ablation}

Using the same screwdriver case and the J4 setting selected in the preceding
subsection, DDPM-generated initialization was compared with structured-random
initialization under identical models, constraints, BO random seed, and
evaluation budget. Each variant used 500 initial candidates followed by 1000
BO evaluations, while Table~\ref{tab:dp_seed_ablation} excludes the initial
candidates from all reported statistics.

DDPM initialization yields 577 all-stage-feasible evaluations and 79 unique
feasible layouts, whereas structured-random initialization yields none. It
also increases the mean feasible-stage ratio from 0.500 to 0.831, reduces the
mean violation degree from 0.386 to 0.100, and raises the trimmed-mean objective
from 0.348 to 0.711. Thus, in this controlled single-seed comparison, DDPM
initialization directed BO toward support regions that remained feasible
across the complete sequence.

\section{Conclusion and Future Work}

This paper presented a modular vacuum-based fixturing system with a cross-stage
planning framework that determines one product-specific shared layout for an
entire predefined robotic disassembly sequence. Physics-based evaluation, DDPM
initialization, and Bayesian optimization jointly determine the balloon-hand
positions and models under stage-dependent geometry and task loads.

The planned layouts were evaluated through physical experiments with a
screwdriver and a shaver. The qualitative vise evaluation showed that rigid fixturing may provide insufficient support for irregular surfaces or obstruct the workspace required for subsequent disassembly. Without
reconfiguring the corresponding shared layout, the robot completed the screw
operations and all 60 component-removal trials, with 30 successful trials for
each object. The 140 screw-removal attempts achieved an aggregate success rate
of 86.4\%, and the observed failures were primarily associated with
screw-alignment errors. Directional external-load tests further revealed
anisotropic and coupled force--moment responses. Comparisons between the
measured directional responses and operational loads yielded mean empirical
stability margins of 66.9\% for the screwdriver and 81.6\% for the shaver.

Together, these results demonstrate that the system can determine a
product-specific shared layout that remains applicable throughout the tested
disassembly sequence. Computational evaluation further showed that the
selected objective concentrated BO in the all-stage-feasible region, while
the controlled ablation showed that DDPM initialization improved the
identification of cross-stage-feasible configurations.

Future work will evaluate products with broader geometries, mass
distributions, and disassembly sequences. Object-pose estimation,
contact-state sensing, and online support adjustment will be investigated to
reduce sim-to-real discrepancies in placement and balloon-hand contact.
Visual detection, pose estimation, and autonomous end-effector alignment will
also replace the manual alignment used in the present screw and
component-removal experiments, advancing an automated workflow from fixture
deployment to robotic disassembly.

\section*{Acknowledgment}
This work was supported by the New Energy and Industrial Technology Development Organization (NEDO) project JPNP23002.

\bibliographystyle{IEEEtran}
\bibliography{paper}

\begin{IEEEbiography}
[{\includegraphics[width=1in,height=1.25in,clip,keepaspectratio]{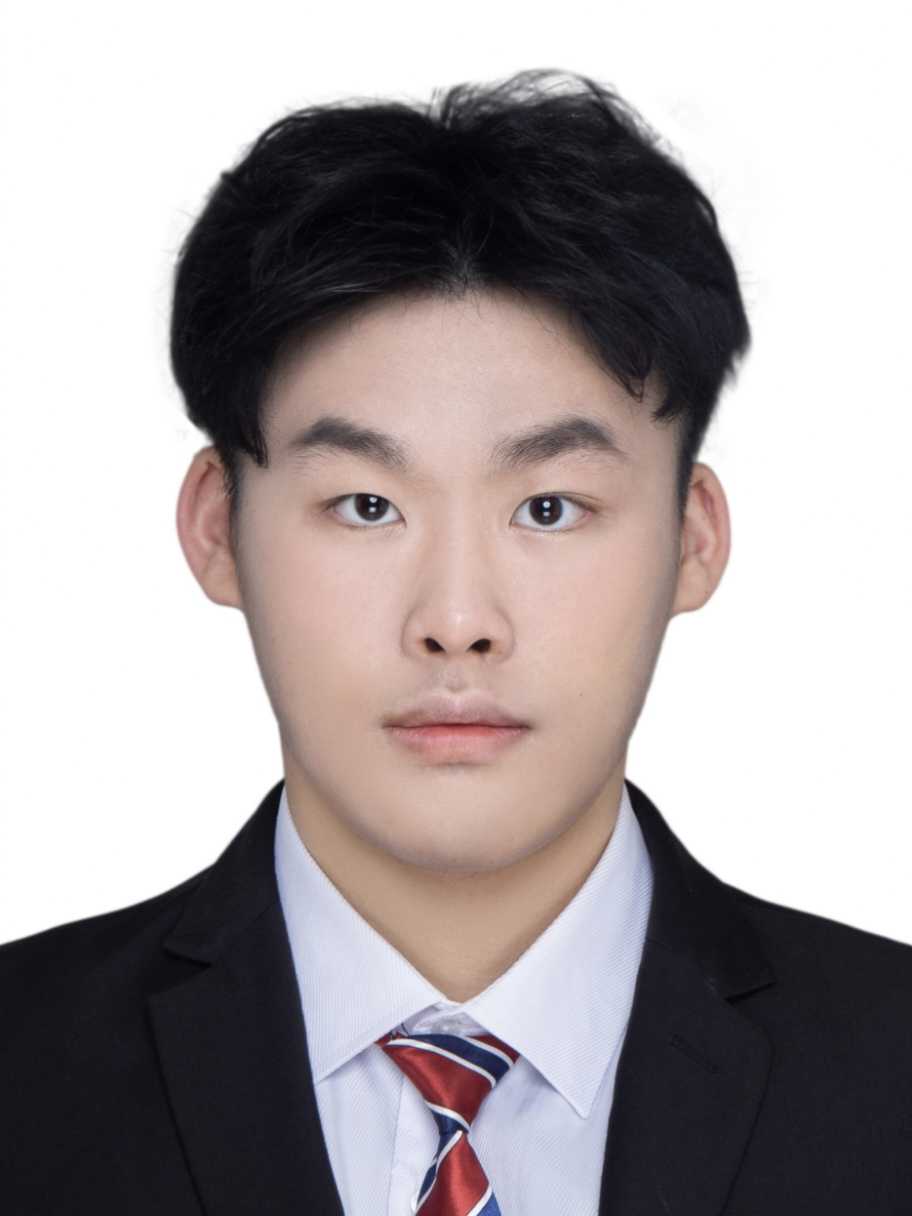}}]
{Haohui Pan} received the B.Eng. degree in robotics engineering from the
College of Mechanical and Vehicle Engineering, Chongqing University,
Chongqing, China, in 2024. He is currently pursuing the M.S. degree with the
Graduate School of Engineering Science, The University of Osaka, Osaka, Japan.
His research interests include robotic disassembly, adaptive fixturing, and
robotic manipulation.
\end{IEEEbiography}

\begin{IEEEbiography}
[{\includegraphics[width=1in,height=1.25in,clip,keepaspectratio]{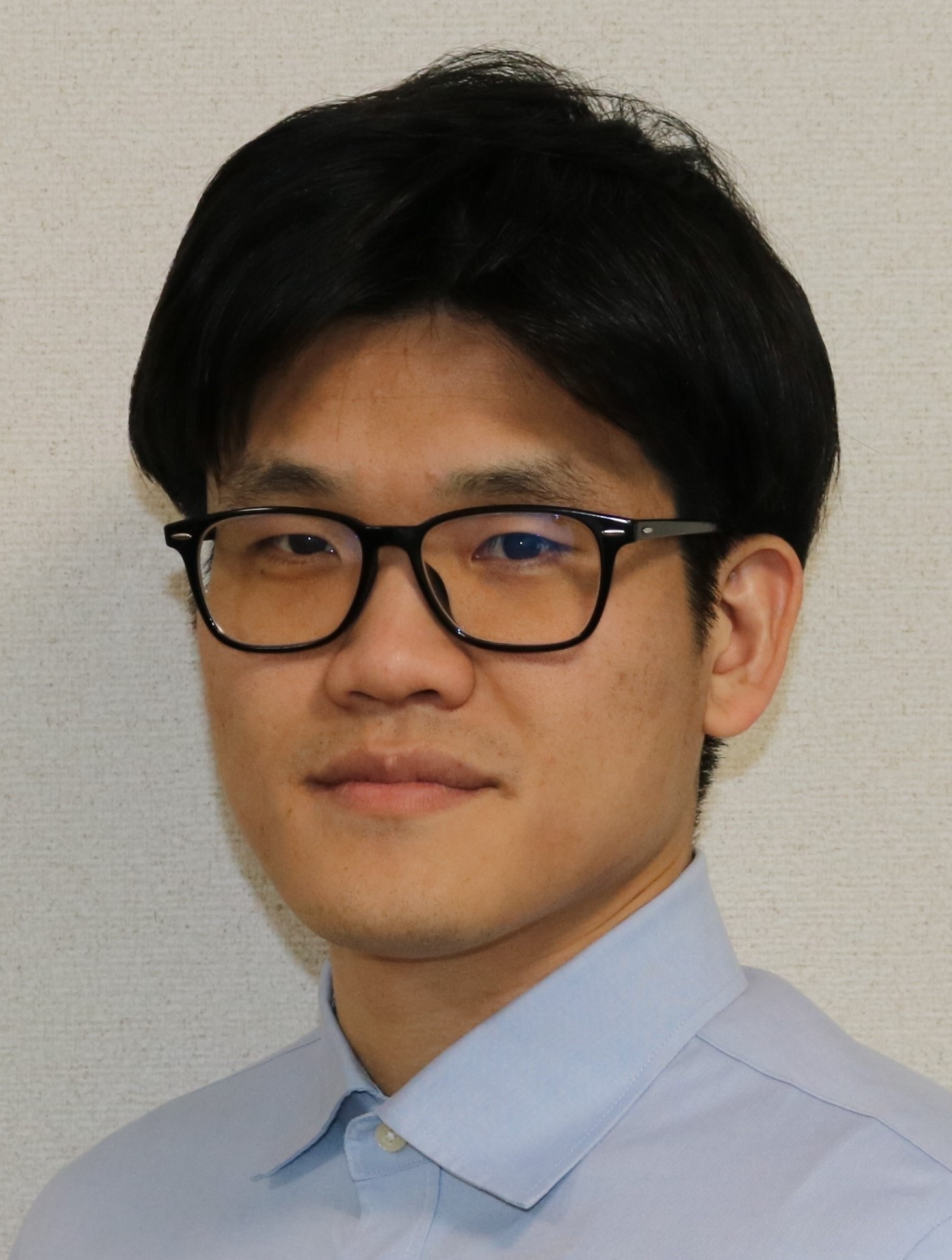}}]
{Takuya Kiyokawa} (Member, IEEE) received his B.E. degree from the National Institute of Technology, Kumamoto College, Japan, and M.E. degree and Ph.D. degree in engineering from the Nara Institute of Science and Technology, Japan, in 2018 and 2021, respectively. From 2021 to 2022, he worked at the University of Osaka, Japan, as a specially appointed Assistant Professor, and at the Nara Institute of Science and Technology, also as a specially appointed Assistant Professor. From 2023 to 2024, he was a Visiting Researcher at the Institute of Robotics and Mechatronics, German Aerospace Center (DLR), Oberpfaffenhofen, Weßling, Germany, and he has been with the University of Osaka as an Assistant Professor since 2023. His current research interests include robotic manipulation and agile reconfigurable robotic systems.
\end{IEEEbiography}

\begin{IEEEbiography}
[{\includegraphics[width=1in,height=1.25in,clip,keepaspectratio]{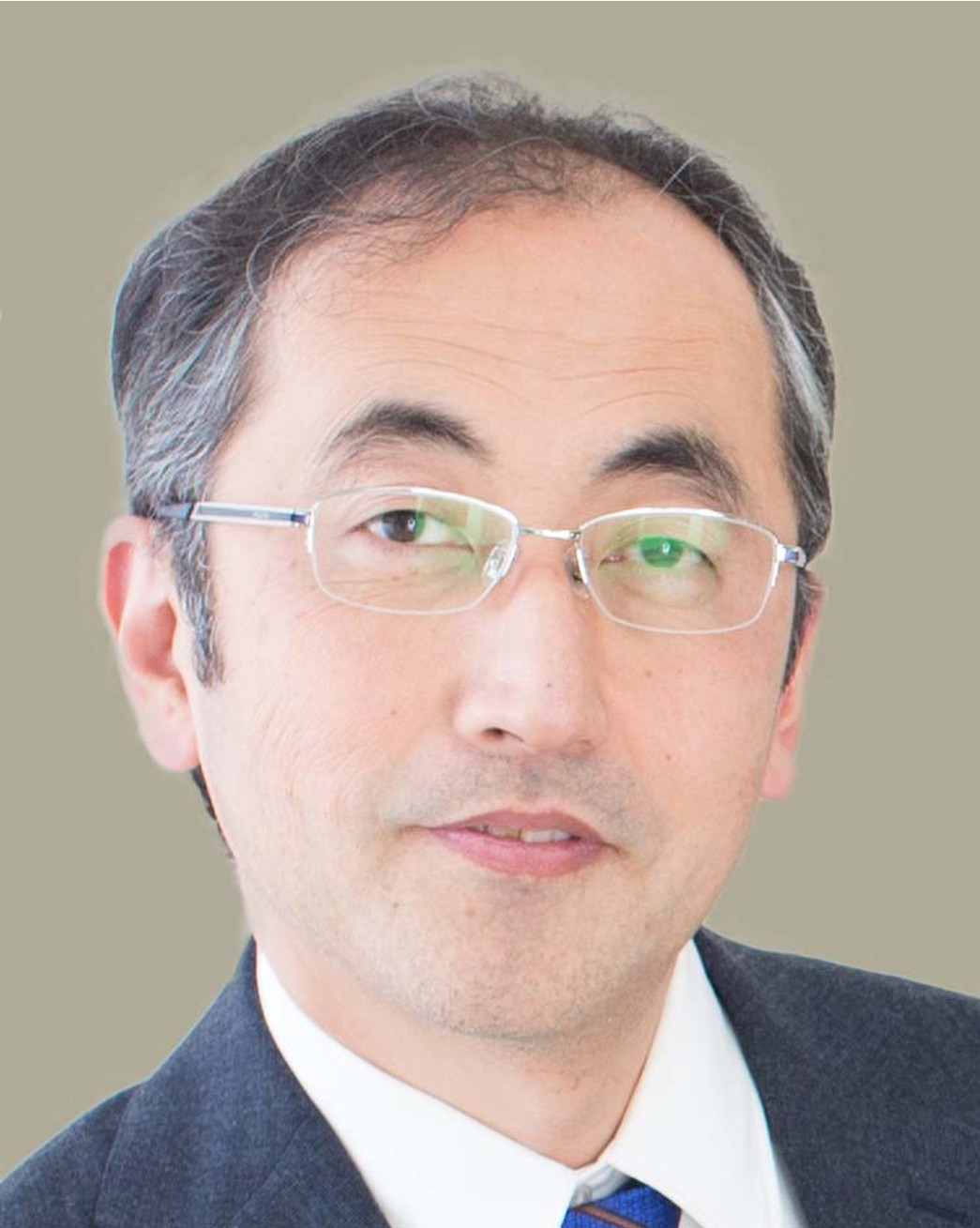}}]
{Kensuke Harada} (Fellow, IEEE) received the Ph.D. degree from the Graduate
School of Mechanical Engineering, Kyoto University, Kyoto, Japan, in 1997. He
is currently a Professor with the Graduate School of Engineering Science, The
University of Osaka, Osaka, Japan. From 1997 to 2002, he was a Research
Associate with the Graduate School of Industrial and Systems Engineering,
Hiroshima University, Hiroshima, Japan. From 2005 to 2006, he was a Visiting
Scholar with the Computer Science Department, Stanford University, Stanford,
CA, USA. Before joining The University of Osaka, he was a Researcher with the
National Institute of Advanced Industrial Science and Technology, Tsukuba,
Japan. His research interests include the mechanics and control of humanoid
robots and robotic hands.
\end{IEEEbiography}

\end{document}